\documentclass[10pt,journal,compsoc]{IEEEtran}
\usepackage{amsmath,amsfonts}
\usepackage{algorithmic}
\usepackage{algorithm}
\usepackage{array} 
\usepackage[caption=false,font=normalsize,labelfont=sf,textfont=sf]{subfig}
\usepackage{textcomp}
\usepackage{stfloats}
\usepackage{url}
\usepackage{verbatim}
\usepackage{graphicx} 
\usepackage{hyperref}
\usepackage{cite}
\usepackage{xcolor}
\usepackage{amssymb}
\usepackage{booktabs}
\usepackage{color}
\usepackage{colortbl}
\usepackage{textcomp}
\usepackage{stfloats}
\usepackage{url}
\usepackage{verbatim}
\usepackage{graphicx}
\usepackage{cite}
\usepackage{marvosym}
\usepackage{booktabs} 
\usepackage{multirow}
\usepackage{tabularx}

\usepackage{hyperref}
\usepackage{url}
\usepackage[utf8]{inputenc} 
\usepackage[T1]{fontenc}    
\usepackage{hyperref}       
\usepackage{url}            
\usepackage{booktabs}       
\usepackage{amsfonts}       
\usepackage{nicefrac}       
\usepackage{microtype}      
\usepackage{xcolor}         
\usepackage{amsmath}
\usepackage{amssymb}
\usepackage{mathtools}
\usepackage{amsthm}
\usepackage{epsfig}
\usepackage{amsfonts}
\usepackage{booktabs}
\usepackage{multirow}
\usepackage{float}
\usepackage{amsmath}
\usepackage{amssymb}
\usepackage{bm}
\usepackage{esint}
\usepackage{colortbl}
\usepackage{wrapfig}
\usepackage{bbm}
\usepackage{multirow}
\definecolor{origin}{rgb}{0.2,0.3,0.9}
\definecolor{mygray}{gray}{.9}
\definecolor{mygray2}{gray}{.8}
\usepackage{hyperref}
\usepackage{url}
\usepackage{graphicx}
\usepackage{amsmath}
\usepackage{mathrsfs}
\usepackage{wrapfig}
\usepackage{booktabs}       
\usepackage{amsfonts}       
\usepackage{nicefrac}       
\usepackage{microtype}      
\usepackage{multirow}
\usepackage{multicol}
\usepackage{makecell}
\usepackage{mathtools}
\usepackage{xspace}
\usepackage{amssymb}
\usepackage{algorithm}
\usepackage{algorithmic}
\usepackage{colortbl}
\usepackage{arydshln}
\usepackage{float}

\definecolor{citecolor}{HTML}{0071BC}
\definecolor{linkcolor}{HTML}{ED1C24}
\definecolor{grey}{HTML}{999999}
\definecolor{green}{HTML}{ABD1BC}
\definecolor{lightblue}{HTML}{B0C4DE}
\definecolor{purple}{HTML}{E3BBED}
\definecolor{orange}{HTML}{ffdab9}
\newlength\savewidth

\theoremstyle{plain}
\definecolor{grass-green}{rgb}{0.4, 0.75, 0.4}
\newtheorem{theorem}{Theorem}[section]
\newtheorem{proposition}[theorem]{Proposition}
\newtheorem{lemma}[theorem]{Lemma}

\theoremstyle{definition}

\theoremstyle{remark}

\definecolor{lightmauve}{rgb}{0.86, 0.82, 1.0}

\begin{document}

\title{Understanding Data Influence with Differential Approximation}

\author{~~~ \\
        Haoru Tan,~~~
        Sitong Wu,~~~ 
        Xiuzhe Wu,~~~ 
        Wang Wang,~~~ 
        Bo Zhao,~~~ 
        Zeke Xie,~~~ \\
        ~~\\
        Gui-Song Xia,~~~
        $\text{Xiaojuan Qi}^{\text{\Letter}}$
\IEEEcompsocitemizethanks{
\IEEEcompsocthanksitem Haoru Tan and Prof. Xiaojuan Qi are with the Department of Electrical and Electronic Engineering at the University of Hong Kong, Pokfulam, Hong Kong. Wang Wang is also with the University of Hong Kong, Pokfulam, Hong Kong. 
\IEEEcompsocthanksitem Sitong Wu is with the Chinese University of Hong Kong, Sha Tin, Hong Kong. 
 \IEEEcompsocthanksitem Prof. Bo Zhao is with Shanghai Jiao Tong University, Minhang District, Shanghai, China. 
 \IEEEcompsocthanksitem Prof. Zeke Xie is with Hong Kong University of Science and Technology (Guangzhou), Nansha District, Guangzhou, Guangdong, China. 
\IEEEcompsocthanksitem Prof. Gui-Song Xia is with the School of Artificial Intelligence at Wuhan University, Wuchang District, Wuhan, Hubei Province, China.  
\IEEEcompsocthanksitem Dr. Xiuzhe Wu is with Stanford University, Stanford, California, USA. 
\IEEEcompsocthanksitem{${\text{\Letter}}$ Corresponding to Prof. Xiaojuan Qi}
}}

\markboth{IEEE Transactions on Pattern Analysis and Machine Intelligence}%
{Shell \MakeLowercase{\textit{et al.}}: A Sample Article Using IEEEtran.cls for IEEE Journals}

\maketitle

\begin{abstract}
Data plays a pivotal role in the groundbreaking advancements in artificial intelligence. The quantitative analysis of data significantly contributes to model training, enhancing both the efficiency and quality of data utilization. However, existing data analysis tools often lag in accuracy. For instance, many of these tools even assume that the loss function of neural networks is convex. These limitations make it challenging to implement current methods effectively. 
In this paper, we introduce a new formulation to approximate a sample's influence by accumulating the differences in influence between consecutive learning steps, which we term Diff-In. Specifically, we formulate the sample-wise influence as the cumulative sum of its changes/differences across successive training iterations. 
By employing second-order approximations, we approximate these difference terms with high accuracy while eliminating the need for model convexity required by existing methods.
Despite being a second-order method, Diff-In maintains computational complexity comparable to that of first-order methods and remains scalable. This efficiency is achieved by computing the product of the Hessian and gradient, which can be efficiently approximated using finite differences of first-order gradients. 
We assess the approximation accuracy of Diff-In both theoretically and empirically. Our theoretical analysis demonstrates that Diff-In achieves significantly lower approximation error compared to existing influence estimators. Extensive experiments further confirm its superior performance across multiple benchmark datasets in three data-centric tasks: data cleaning, data deletion, and coreset selection. 
Notably, our experiments on data pruning for large-scale vision-language pre-training show that Diff-In can scale to millions of data points and outperforms strong baselines. 
\end{abstract}


\section{Introduction}
\label{sec:1}

Data is a driving force behind recent advancements in various fields \cite{GPT3, SAM}, as it directly influences the behavior of learned models, including their performance and inherent biases \cite{kwon2023datainf}. This highlights the need for a quantitative understanding of how individual data samples affect model learning, which is essential for enhancing both model performance \cite{OPT,xia2024less} and interpretability \cite{Studying_large_language,Attribution_diffusion}. To address this, influence functions have been introduced \cite{alma991001947589706011,cook1986assessment} to study how a specific sample
$z$ affects model parameters and loss values:
\begin{equation}
\begin{aligned}
\label{eq: original influence on params}
    &\mathcal{I}_{\theta}({z}) =\theta^{*}_{-z} - \theta^{*}, \quad\quad\quad\quad\quad~~\text{(Influence on parameters)} \\
    &\mathcal{I}({z}, \mathbf{V}) = \mathcal{L}(\mathbf{V}, \theta^{*}_{-z}) - \mathcal{L}(\mathbf{V}, \theta^{*}).\quad\quad \text{(Influence on loss)}
 \end{aligned} 
\end{equation}
Here, $\theta^{*}$ represents the learned parameters obtained by optimizing the empirical loss $\mathcal{L}$ on the full training set, while $\theta^{*}_{-z}$ refers to the parameters learned after excluding the sample $z$. The influence on parameters,  $\mathcal{I}_{\theta}({z})$, also known as Cook's distance \cite{alma991001947589706011,cook1986assessment}, measures the extent to which the optimized model parameters would change if the sample $z$ is removed from the training dataset.  
Similarly, the influence on loss, $\mathcal{I}({z}, \mathbf{V})$, examines how the model's loss or performance on an evaluation set $\mathbf{V}$ is affected when the sample $z$ is excluded from the training set.

\begin{figure*}[tp] 
\centering
\hspace{-0.328cm}\includegraphics[width=0.9999188\linewidth]{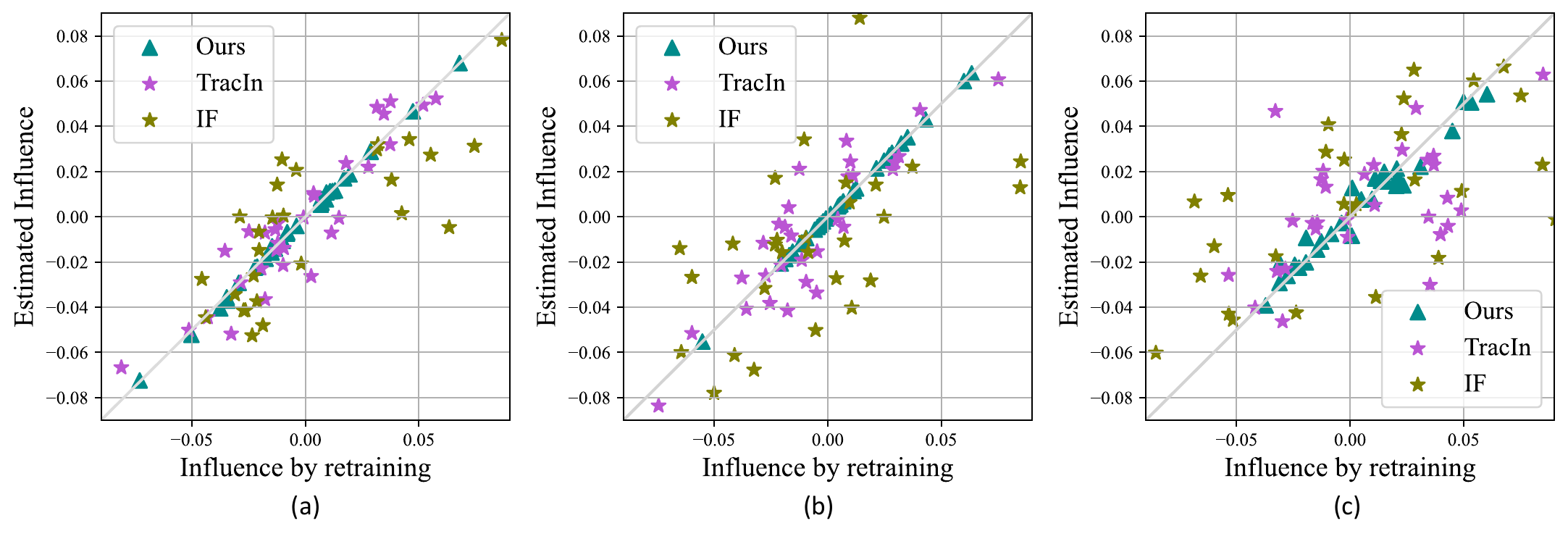}
\vspace{-0.5cm}
\caption{\label{fig: scatter comp}
Approximation accuracy comparison by comparing the estimated influence values with actual influence values obtained through brute-force retraining on the 30 most influential data points. There are three models and dataset settings: (a) ResNet-18 on CIFAR-10, (b) ResNet-101 on CIFAR-10, and (c) ResNet-18 on ImageNet-1K. 
The more accurate a method is, the closer its corresponding scatter will be concentrated near the diagonal. 
Our approach demonstrates consistent advantages as the model size grows (a vs. b) and as the dataset complexity increases (a vs. c). 
}  
\end{figure*}

To measure the influence of a sample $z$, a straightforward yet optimal approach would be to remove $z$ from the training dataset and retrain the model to obtain the optimized parameters $\theta^{*}_{-z}$, a process known as leave-one-out (LOO) training. However, the retraining is computationally expensive and often impractical. To overcome this limitation, Koh and Liang \cite{IF}, building on the formulation of influence functions \cite{alma991001947589706011}, introduced approximations that estimate a sample's influence without the need for retraining. 
The key idea is to derive a quadratic approximation of the empirical risk around the stationary point $\theta^{*}$ and {obtain the sample's influence by upweighting it by an infinitesimal amount}; see Sec. \ref{sec: Representative Methods} for details. 
Subsequent studies \cite{group_rffect_improve,group_rffect_analysis,fragile,Studying_large_language,kwon2023datainf,Mirrored} have expanded on this approach, improving both its efficiency and precision. 
However, despite these advancements, the accuracy of these methods relies on the convexity \cite{IF} of the model, conditions that are rarely satisfied in practice, especially in large models. As a result, these approximations can be inaccurate, as demonstrated in Figure \ref{fig: scatter comp} where the sample's influence (olive stars) deviates significantly from the LOO results, showing a low correlation. This limitation also impacts their performance in practice (see Sec. \ref{sec:experiment}).

Recently, another line of work \cite{TracIN,wang2025capturingDVE,tan2023data,hara2019data} accumulates gradient information over the training trajectory to quantify each sample’s contribution; a notable example is TracIn \cite{TracIN}, which offers a heuristic method for approximating a sample $z$'s influence on loss values, requiring only gradient computations. This method accumulates the sample's impact on the validation set loss across various training iterations through first-order approximations of the loss. Although TracIn is more computationally efficient, its approximation diverges significantly from the objective defined in Eq.\eqref{eq: original influence on params}, limiting its performance, see Sec.\ref{sec:experiment}. Additionally, it cannot be used to estimate a sample's influence on model parameters. Other approaches, such as SGD‑INF \cite{hara2019data} are difficult to apply because of their large computational overhead; DVE‑INF \cite{wang2025capturingDVE}, although significantly accelerated by random‑projection strategies, incurs a non‑negligible loss in estimation accuracy (see Table \ref{tb: approximation accuracy test}).

In this paper, we introduce a new perspective on influence estimation by examining its temporal differences, termed Diff-In. The core of Diff-In is to represent influence as the cumulative sum of differences between successive training steps (see Eq.\eqref{eq: differentiate}). Although simple, this formulation allows the application of a second-order approximation to each difference term without convexity assumptions \cite{IF} or altering the approximation target \cite{TracIN}, thereby enhancing accuracy (see Figure \ref{fig: scatter comp} green triangle). This improvement is demonstrated both theoretically (Sec. \ref{sec:theoretical}) and empirically (Sec. \ref{subsec: Preliminary experiments}).
Moreover, although Diff-In employs a second-order approximation, it does not significantly increase computational complexity compared to existing methods \cite{IF,TracIN} (see Sec. \ref{subsec: Preliminary experiments}). Instead of directly computing second-order derivatives, Diff-In calculates the product of the Hessian and the gradient \cite{fast_hessian}. This is done using finite differences on the gradient, as shown in Eq.\eqref{eq: fast hessian}, requiring only gradient computations and maintaining an efficiency with previous first-order methods \cite{TracIN}. We provide a functional comparison of our Diff-In method with representative prior works in Table \ref{tb: method comp}.

\begin{table*}[tp]
\setlength{\tabcolsep}{3.1pt}
\caption{\label{tb: method comp} 
We compare several representative influence estimators. Both IF \cite{IF} and DataInf \cite{kwon2023datainf} require the assumptions of loss convexity and stationarity (i.e., the solution should be a stationary point). GEX \cite{GEX}, TracIn \cite{TracIN}, and Forward-INF \cite{Mirrored} cannot estimate the influence of samples on parameters, as defined in Eq.\eqref{eq: original influence on params}. In contrast, while SGD-INF \cite{hara2019data} and DVE-INF \cite{wang2025capturingDVE} achieve similar outcomes as our Diff-In method across various entries, their accuracy is inferior to ours, both theoretically (see Sec. \ref{sec:theoretical}) and empirically (see Sec. \ref{sec:experiment}). 
}
\centering
\resizebox{0.988\linewidth}{!}{ 
\begin{tabular}{lccccccccccccc}
    \toprule[1.3pt]
    {~} & IF series \cite{IF,Studying_large_language,kwon2023datainf,Arnoldi,group_rffect_analysis} & ~~~~~GEX \cite{GEX}~~~~~  & ~~TracIn \cite{TracIN}~~ & ~~Forward-INF \cite{Mirrored}~~  & ~~SGD-INF \cite{hara2019data}~~ & ~~DVE-INF \cite{wang2025capturingDVE}~~ & ~~Diff-In (Ours)~~\\
    \midrule[1.3pt]
    {Avoids loss convexity}  & $\times$ & $\checkmark$ & $\checkmark$ & $\checkmark$ & $\checkmark$ & $\checkmark$ & $\checkmark$\\
    \midrule[0.4pt]
    {Avoids stationarity} & $\times$ & $\checkmark$ & $\checkmark$ & $\checkmark$ & $\checkmark$ & $\checkmark$ & $\checkmark$\\
    \midrule[0.4pt]
    {Leave-one-out influence}  &$\checkmark$ & $\checkmark$ &  $\times$ & $\checkmark$ & $\checkmark$ & $\checkmark$ & $\checkmark$\\
    \midrule[0.4pt]
    {Training dynamics} & $\times$ & $\times$ & $\checkmark$ & $\checkmark$ & $\checkmark$ & $\checkmark$ & $\checkmark$ \\
    \midrule[0.4pt]
    {Influence on loss} & $\checkmark$ & $\checkmark$ & $\checkmark$ & $\checkmark$ & $\checkmark$ & $\checkmark$ & $\checkmark$\\
    \midrule[0.4pt]
    {Influence on parameters}  & $\checkmark$  & $\times$ & $\times$ & $\times$ & $\checkmark$ & $\checkmark$ & $\checkmark$\\
\bottomrule[1.3pt]
\end{tabular}
}
\end{table*}

We conduct extensive experiments on various data-centric tasks, including coreset selection \cite{SSP}, data cleaning \cite{TracIN}, and data deletion \cite{unlearning_iclr}. The results demonstrate that Diff-In consistently outperforms previous methods \cite{IF, TracIN, kwon2023datainf} across all tasks and datasets, delivering leading performance in most evaluated scenarios. Thanks to its precise estimation, Diff‑In achieves leading performances across multiple empirical benchmarks. For data‑cleaning on image classification (including ImageNet) and LLM-related tasks, Diff-In outperforms the prior best method by 6\% to 10\%. In data deletion experiments, Diff-In can largely eliminate the harmful effects of noisy training samples without retraining. For example, given an LLM (Llama 3.1 8B \cite{dubey2024llama3herdmodels}) on corrupted GSM8K \cite{cobbe2021gsm8k}, applying Diff-In improves model performance by 8.2\% over the corrupted baseline; this result is within 2.4 percentage points of brute‑force retraining and surpasses other data‑deletion methods by 4.2 percentage points. 
In coreset‑selection experiments, using Diff‑In as an indicator for selecting multimodal pretraining data yields consistent downstream gains. For example, with a 60\% selection budget, Diff‑In outperforms the popular CLIP‑score selection method by 1.3–2.1 percentage points on downstream tasks such as zero‑shot and linear‑probe image classification. 
Additionally, our approach is versatile, effectively solving all three tasks, while some of the compared methods \cite{Moderate,TracIN,GEX} are tailored to specific tasks and cannot be applied across different data-centric tasks.

\section{Related Works}

\label{sec: detailed related works}

\subsection{Influence Analysis}
\label{sec: related work influence analysis}

Influence analysis is a technique to elucidate the connection between training data and model predictions \cite{survey, cook1986assessment, IF, shapley1953value, kernel_shapley}. A straightforward way to measure the significance of a sample is to perform leave-one-out (LOO) retraining, which involves retraining the model on the training set without the sample and then comparing the discrepancy between it and the model trained on the full set. Nevertheless, the high computational expense of retraining for each sample is generally prohibitive due to the enormous size of contemporary deep-learning datasets \cite{imagenet,LAION,CC12M}. Even some works that try to reduce the cost of retraining, such as by dividing the full set into subsets \cite{downsampling}, still incur a high overall expense.

Instead of performing any retraining, Koh et al. \cite{IF} estimate the model change from a small weight change of training data. Specifically, this method utilizes the inverse-Hessian-gradient-product-based estimator to approximate the sample-wise influence. 
However, Koh's estimator has some limitations. 
The first one is that it relies on an overly strong assumption that the loss function with parameters should be convex, which is often not the case \cite{choromanska2015loss, dauphin2014identifying}. 
The second limitation is that it can not be easily scaled up to large models and large datasets due to the heavy computation of the inverse-Hessian-gradient-product. 
The third shortcoming is that it cannot model the training dynamics since it only uses the parameters checkpoint from the very last iteration. 
Even so, it has attracted considerable attention from the academic community. 
Subsequently, there have been various research efforts to improve this method in various directions. 
For better scalability, a line of work \cite{Studying_large_language, Arnoldi, kwon2023datainf} has been proposed. 
\cite{Studying_large_language} proposed an efficient decomposition of the Hessian to speed up the estimator's calculation. 
Another excellent work \cite{Arnoldi} introduced the novel Arnoldi iteration technique for accelerating the computation of the inverse Hessian, enabling applications to large-scale Transformer models in language and vision tasks, even when they have hundreds of millions of parameters. 
DataInf \cite{kwon2023datainf} proposed a novel closed-form approximation for the inverse-Hessian with better efficiency in both computational and memory
complexities. 
{\cite{klochkov2024revisitinginversehessianvector} studied some hyperparameters affecting the precision of using the LiSSA algorithm to calculate the Hessian-related operation for the influence function.
\cite{IP} proposed a Hessian-free approach to estimate the influence function by only calculating the inner product in the gradient space, thereby achieving better scalability.} 
As for the group effect \cite{OPT}, some works also analyze \cite{group_rffect_analysis} and improve \cite{group_rffect_improve} the influence function on measuring group effects, for instance, \cite{group_rffect_improve} extended influence functions to directly account for sub-population-group effects by considering higher-order terms in Taylor approximation. 
Moreover, PBRF \cite{PBRF} analyzes several practical reasons for the failure of Koh's estimator, \textit{e.g.} the distinction between cold-start and warm-start response, the implicit regularizer, and the non-converged parameters. Additionally, it proposes the proximal Bregman response function to improve the performance. 
However, the requirement of Koh's estimator on the convex property of the loss function and the neglect of training dynamics are not well solved in the above method. 
Recently, some work has attempted to estimate the effect of a sample by comparing the changes in the representation of some samples before and after additional training \cite{GEX,Mirrored}.
They will perform better than Koh's estimator, but additional training will undoubtedly bring additional computational costs, especially if the data set is large. 
\cite{chhabra2024what} proposed to combine the influence function with tree structure to provide interpretations of which sample features contribute positively or negatively to the model’s performance.

In recent years, a series of works represented by SGD-Inf \cite{hara2019data}, TracIn \cite{TracIN}, and MoSo \cite{tan2023data} have brought new solutions to this problem. All three are achieved by calculating the gradient information during training. For example, SGD-Inf \cite{hara2019data} proposes an approximator with theoretical guarantees by tracking the gradient information of each sample at every epoch during training; however, it is highly time-consuming. In contrast, DVE-INF \cite{wang2025capturingDVE} significantly improves computational efficiency by utilizing time step sampling and random projection techniques. These schemes not only eliminate the requirement of convexity of the loss function in the calculation of influence but also fully perceive training dynamics. 
However, these methods are also problematic, as they either do not estimate the influence of a sample on model parameters. In addition, in training, the choice of optimizer also has a significant impact on training, and these schemes are designed for one particular optimizer, namely, the stochastic gradient descent (SGD). So they don't account for the impact of the optimizer's variability (\textit{e.g.} the widely used Adam \cite{adam}), thereby limiting their broader applicability.

Besides the aforementioned methods, another very effective scheme for estimating the influence is the Shapley value \cite{shapley1953value} based on cooperative game theory. It can be interpreted as the contribution of each sample to the model's performance by quantifying the marginal gain in performance when a sample is added to a randomly selected subset. However, the computation of the Shapley value is very expensive, since it requires evaluating all possible subsets. Therefore, some approximation methods, such as Monte-Carlo sampling \cite{TMC}, kernel method \cite{kernel_shapley}, or KNN-based method \cite{KNNshap}, have been proposed to reduce the computational cost.

\subsection{Applications} 
\label{sec: related work applications}

The technique of influence analysis, which measures the impact of training samples on the model's performance, has been explored by the academic community in many scenarios. Here we list some representative topics and works. (1) Dataset pruning / Coreset selection \cite{OPT,knn-loo,choe2024dataworthgptllmscale}: selecting a subset of the dataset that preserves the model's accuracy while reducing the size or complexity of the data. 
{For example, LoGra \cite{choe2024dataworthgptllmscale} proposed an extremely efficient influence-based pipeline to conduct data valuation and selection for large language models.}  
(2) Noise and {outlier sample detection \cite{IF,knn-loo,Shapley_noise_clean_zou,hara2019data, OGI}} and noise label correction \cite{data_correction}: identifying and removing or correcting the samples that have incorrect or misleading labels or features that degrade the model's quality. (3) Adversarial attack \cite{IF,attack_poisoning}: generating samples that can fool or attack the model by exploiting its weaknesses or vulnerabilities. (4) Continual learning \cite{continual_learning}: the process of constantly monitoring and retraining machine learning models with updated data to prevent concept drifts and maintain accuracy and reliability. (5) Machine unlearning \cite{unlearning_iclr}: removing the influence of a specific sample or group of samples from the model, which can be useful for privacy, security, or legal reasons. (6) Data attribution \cite{Studying_large_language,Explaining_a_series,Attribution_acale,Attribution_diffusion,ilyas2022datamodels}: attributing the model's output or behavior to the input data or features that contributed to it. For example, \cite{li2022achievingfairnessutilitycost} proposed an influence-based data reweighting pipeline to enhance fairness. HYDRA \cite{chen2022hydrahypergradientdatarelevance} attributes the model's output by unrolling the hyper-gradient of test loss throughout the training trajectory. 
Datamodels \cite{ilyas2022datamodels} proposed an interpretable pipeline by introducing a simple surrogate model (like a linear model) to understand the relation between data and prediction, and then give rise to various interesting applications. (7) ISAL \cite{ISAL} designed an active learning pipeline by utilizing the influence function to pick up the most influential data points at each iteration. 

\begin{table*}[tp]  
\centering
\caption{Comprehensive overview of the notational convention. \label{tb:notation}}
\begin{tabularx}{0.98\linewidth}{lX}
\toprule
Notation~~\quad~~~~~~~&Description\\
\midrule
$\mathbf{D}$        & The training set, and the size of the training set is $|\mathbf{D}|=N$.\\
$\langle \cdot, \cdot \rangle$        & The inner-product operator.\\
$z \in \mathbf{D}$        & A training data.\\
$\mathbf{D}/z$        & The training set excluded the sample $z$.\\
$\mathbf{V}$        & The validation set.\\
$\mathbf{B}_t$        & The mini-batch at the $t$-th iteration.\\
$\ell(\cdot)$                 & The loss function over one data point.\\
$\mathcal{L}(\cdot)$                 & The averaged loss over batch or set.\\
$\theta^*$                 & The learned parameters optimized on the full training set after training.\\
$\theta^t$                 & The learned parameters optimized on the full training set at the $t$-th iteration.\\
$\theta^*_{-z}$                 & The learned parameters optimized on the dataset excluded the sample $z$  after training.\\
$\theta^t_{-z}$                 & The learned parameters optimized on the dataset excluded the sample $z$ at the $t$-th iteration.\\
$\theta^*_{-\mathbf{Z}}$                 & The learned parameters optimized on the dataset excluded a sample set $\mathbf{Z}$  after training.\\
$p$                 & The number of trainable parameters.\\
$G$                 & The gradient of the model parameter.\\
$H$                 & The Hessian of the model parameter.\\
$G^t_z$                 & The gradient of the parameters at the $t$-th training iteration over the sample $z$.\\
$H^t_z$                 & The Hessian of the parameters at the $t$-th training iteration over the sample $z$.\\
$G^t_{-z}$                 & The gradient of the parameters at the $t$-th training iteration over $\mathbf{D}/z$.\\
$T$                 & The maximum iteration of the training process.\\
$\mathcal{T}_m$                 & A set of $m$ uniformly selected time-steps $\mathcal{T}_m=\{t_1,...,t_m\}$.\\
$m$                 &The number of uniformly selected time-steps in $\mathcal{T}_m$.\\
$\mathcal{I}_\theta(z),$                 & The influences on parameters, also the Cook's distance \cite{cook1986assessment}.\\
$\mathcal{I}(z,\mathbf{V})$                 & The influences on loss over the validation set $\mathbf{V}$.\\
$\mathcal{D}^t$                 & The influence difference between two adjacent time steps $t$ and $t-1$.\\
$\eta$                 & The learning rate in the optimizer.\\
$\beta$                 & The momentum weight in the optimizer.\\
$\ell$                 & The Lipschitz constant.\\
$g$                 & The upper bound of the gradient norm.\\
$C$                 & The farthest distance the neural network parameters move away from their initial state during training when any subset $\mathbf{D}_s \subset \mathbf{D}$ is used as the training set. \\
LOO                 & Leave-one-out. \\
\bottomrule 
\end{tabularx}
\end{table*}

\section{Preliminaries: Revisiting Influence Estimation}
\label{sec: Representative Methods} 

Table \ref{tb:notation} contains all notations. Let $\mathbf{D} = \{{z}_0, ..., {z}_{N-1}\}$ denote the training set, where the number of all samples $|\mathbf{D}| = N$. 
$\mathbf{D}/z$ is the dataset excluding a sample $z$. 
For a deep network parameterized by $\theta \in \Theta$, we use $\ell({z},\theta)$ as the loss on a sample ${z}$ and $\mathcal{L}(:,\theta)$ as the averaged loss over a set of data, where $\mathcal{L}(\mathbf{D},\theta) = \frac{1}{N}\sum_i \ell({z}_i,\theta)$. 
Please check Eq.\eqref{eq: original influence on params} for the definition of the influence. Here, we review several representative works from previous researchers and discuss their characteristics and limitations. We put the functional comparison between previous methods and our Diff-In in Table \ref{tb: method comp}.

\vspace{0.3cm}
\noindent\textbf{IF-series methods:} 
To mitigate the high cost of brute-force leave-one-out (LOO) retraining, various methods have been proposed to estimate influence \cite{IF, Studying_large_language, kwon2023datainf, TracIN, group_rffect_improve}. A notable example is the approach introduced by Koh and Liang \cite{IF}, which calculates the change in model parameters when a sample $z$ is up-weighted by a small amount $\epsilon$. Specifically, the optimal parameters $\theta^{*}_{\epsilon, z}$, resulting from up-weighting the sample $z$ by $\epsilon$, are formulated as: $\theta^{*}_{\epsilon, z} = 
\arg\min_{\theta\in\Theta} \frac{1}{n} \sum_{i=1}^n \ell(z_i,\theta)
+ \epsilon \ell(z, \theta)$.

Then, according to \cite{alma991001947589706011}, by applying a quadratic approximation to the empirical risk around $\theta^{*}$, the influence of up-weighting $z$ on the parameters $\theta^{}_{\epsilon=0, z} = \theta^{}$ by $\epsilon$ is given by: 
\begin{align}
\label{eqn:params}
\mathcal{I}_\text{up,params}(z) & = \frac{d\theta_{\epsilon, z}^{*}}{d\epsilon}\Bigr|_{\epsilon = 0} = -H_{\theta^{*}}^{-1} \ \nabla_\theta \ell(z, \theta^{*}),
\end{align}
where $H_{\theta^{*}} = \frac{1}{n} \sum_{i=1}^n \nabla^2_\theta \ell(z_i, \theta^{*})$ is the Hessian and is positive definite (PD) by assumption.
 Since removing a point $z$ is the same as up-weighting it by $\epsilon = -\frac{1}{n}$, one can then approximate the parameter by computing
$\mathcal{I}_{\theta}(z)=\theta_{-z}^{*} - \theta^{*} \approx -\frac{1}{n} \mathcal{I}_\text{up,params}(z)$. 

With $\mathcal{I}_\theta(z)$, the influence on the loss over the validation set $\mathbf{V}$ can be estimated as:
$\mathcal{I}({z}, \mathbf{V}) =  \langle \nabla \mathcal{L}(\mathbf{V}, \theta^*),~  \mathcal{I}_{\theta}({z})\rangle$. 
While these methods represent significant progress, they rely on the assumption that the empirical risk is strongly convex in the parameters, an assumption that is rarely satisfied in practice \cite{choromanska2015loss, dauphin2014identifying}. This limitation leads to reduced approximation accuracy (see Figure \ref{fig: scatter comp}) and suboptimal performance in real-world applications (see Sec. \ref{sec:experiment}). Additionally, the need to compute the inverse Hessian constrains the scalability of these methods for large-scale applications.

\vspace{0.3cm}
\noindent\textbf{Gradient accumulation methods:} Recently, another line of work \cite{TracIN,wang2025capturingDVE,tan2023data,hara2019data} accumulates gradient information over the training trajectory to quantify each sample’s contribution; a notable example is TracIn \cite{TracIN, xia2024less}, bypassed the convex loss assumption by introducing a heuristic proxy metric to the original influence metric defined in Eq.\eqref{eq: original influence on params}, that is,  $$\texttt{TracIn-Ideal}(z, \mathbf{V}) = \sum_{t: z_t = z} \mathcal{L}(\mathbf{V}, \theta^{t}) - \mathcal{L}(\mathbf{V}, \theta^{t+1}),$$ 
which metric measures the total reduction in loss on the validation set $\mathbf{V}$ caused by the stochastic gradient descent process whenever the training example $z$ is used. TracIn approximates this heuristic proxy with a first-order estimator:  
\begin{equation}
    \texttt{TracIn}({z}, \mathbf{V}) = \sum_{t \in \mathcal{T}_m} \eta_t \langle \nabla \mathcal{L}(\mathbf{V}, \theta^t),~ \nabla \mathcal{L}(\mathbf{D}, \theta^t) \rangle,
    \vspace{-0.30cm}
\end{equation} 
where $\langle \cdot, \cdot \rangle$ denotes the inner-product operation, $\eta_t$ is the learning rate at the $t$-th iteration, and $\mathcal{T}_m = {t_1, ..., t_m}$ is a set of sampled time steps. Due to its first-order approximation, TracIn is more scalable for large-scale datasets than the method proposed by Koh and Liang \cite{IF}. However, its heuristic goal differs from the original definition, and when combined with approximation errors, it often leads to imprecise estimates \cite{kwon2023datainf}. As shown in Fig. \ref{fig: scatter comp}, the approximation accuracy of TracIn decreases as the dataset size and the number of model parameters increase. Furthermore, this approach is not suitable for estimating a sample $z$'s influence on the parameters. SGD‑INF \cite{hara2019data} and its follow‑on acceleration work \cite{wang2025capturingDVE} account for higher‑order information from the training trajectory, but their computations are extremely complex and require access to every training step. DVE‑INF, while achieving scalability improvements through techniques such as random projection, incurs a non‑negligible drop in estimation accuracy, see Table \ref{tb: approximation accuracy test}.

\section{Influence Estimation via Differential Approximation}
\label{sec: method}

In the following sections, we first explain how to expand the influence as the cumulative sum of differences in Sec. \ref{subsec:Influence differentiation} and introduce an efficient second-order estimator for the difference term (\textbf{Lemma} \ref{lemma 1}) along with its first-order approximations (Eq.\eqref{eq: fast hessian}).
Next, building on the derived difference term, we present Diff-In, which estimates the influence on parameters and loss values in Proposition \ref{tm 1}. 
Finally, we discuss implementation details in Sec. \ref{sec: Implementation Details}.
We will extend the method to other optimizers in Sec. \ref{subsec: How to Extend Diff-In to Other Optimizers?}.

\subsection{Diff-In}
\label{subsec:Influence differentiation}

Here, we formulate the influence as the cumulative sum of its differences between successive training time steps. 
We denote the influence of sample $z$ on the parameters at the $t$-th training iteration as $\mathcal{I}_\theta^t(z) = \theta^{t}_{-z} - \theta^t$. 
The sample-wise influence difference between adjacent training steps is given by $\mathcal{D}^t({z}) = \mathcal{I}_\theta^{t+1}({z}) - \mathcal{I}_\theta^{t}({z})$. Given that the maximum number of iterations is $T$, we have $\mathcal{I}_\theta = \mathcal{I}_\theta^T$. Therefore, we can express $\mathcal{I}_{\theta}^t$ as follows: 
\begin{equation}
\small
\mathcal{I}_\theta({z}) = \sum_{t<T} \Big(\mathcal{I}_\theta^{t+1}(z) - \mathcal{I}_\theta^{t}(z) \Big) + \mathcal{I}_\theta^{0}({z}) = \sum_{t<T} \mathcal{D}^{t}({z}) + \mathcal{I}_\theta^{0}({z}),
\label{eq: differentiate}
\end{equation}
where $\mathcal{I}_\theta^0(z)=\theta^0_{-z} - \theta^0 = 0$, since the removal or inclusion of sample $z$ does not affect the initial model parameters.  
Although simple, this approach allows us to approximate the difference terms without this approach enables us to approximate the difference terms without relying on convexity assumptions, as demonstrated below, thereby enhancing approximation accuracy.  

\vspace{0.3cm}
\subsubsection{Estimation for the difference term $\mathcal{D}^t({z})$}  
We can express $\mathcal{D}^t(z)$ as: $\mathcal{D}^t(z) = (\theta_{-z}^{t+1} - \theta_{-z}^t) - (\theta^{t+1} - \theta^t)$. Assuming the use of the SGD optimizer, according to the parameter update rule, we have  $\theta^{t+1} - \theta^t = -\eta_t G^t$, where $G^t$ represents the gradient at iteration $t$ and $\eta_t$ denotes the learning rate. We can then approximate $\mathcal{D}^t(z) = -\eta_t(G_{-z}^t - G^t)$ where $G_{-z}^t$ denotes the gradient at iteration $t$ with sample $z$  removed from the training set. 
It is important to highlight that our approximation can be readily extended to other optimization methods by adjusting the parameter update rule as necessary (see Sec. \ref{subsec: How to Extend Diff-In to Other Optimizers?}).
By further expressing $G_{-z}^t - G^t$ using a continuous time approximation and introducing a perturbation parameter $\epsilon$ for the gradient difference terms, following \cite{IF},  we can derive $\mathcal{D}^t(z)$ using second-order and first-order terms through Taylor's expansion. 
This leads us to the following Lemma.

\begin{lemma}
\label{lemma 1}
Given the parameters $\theta_D^t$ optimized via the SGD optimizer at the $t$-th iteration, by supposing the time-step to be continuous, $\mathcal{D}^t({z})$ can be approximated as follow, 
\begin{equation}
\hat{\mathcal{D}}^t({z})= \sum_{k \leq t} a_{t,k} \Big( H_{B^k}^k G_z^k + H_z^k ~ G_{B^k}^k \Big),  
\label{eq: further_derivation} 
\end{equation}
where $B^t \subset \mathbf{D}$ is the training mini-batch at the $t$-th,    
$H^t_{B^t} = \nabla^2 \mathcal{L}(B^t, \theta^t)$ is the hessian over the batch, $G^t_{B^t} = \nabla \mathcal{L}(B^t, \theta^t)$ is the gradient over the batch. $H^t_z = \nabla^2 \ell(z, \theta^t)$ and $G^t_z = \nabla \ell(z, \theta^t)$ are the hessian and the gradient on the sample $z$, respectively. 
The coefficient $a_{t,k}={-(\eta_t \eta_k)^2}/{N}$ is a function of the learning rate $\eta_t$ and $\eta_k$. 
\end{lemma} 
The detailed proof is provided in the supplementary material. 
Despite being a second-order estimator, the Hessian-gradient product can be efficiently approximated using the classic finite difference method  \cite{fast_hessian}, as described by the following rule:
\begin{equation}
H G \approx \lim_{\epsilon \rightarrow 0}~ \frac{\nabla \mathcal{L}(\theta + \epsilon G) - \nabla \mathcal{L}(\theta) }{\epsilon},
    \label{eq: fast hessian}
\end{equation}
where the complexity is $\mathcal{O}(p)$, with $p$ representing the number of parameters, making it comparable to first-order methods. This approximation can be easily implemented and efficiently executed using existing deep learning frameworks like PyTorch \cite{pytorch}.

\vspace{0.3cm}
\subsubsection{Influence on parameters and loss} Using the approximation for the difference term of influence on parameters, $\hat{\mathcal{D}}^t({z})$, as shown in Eq.\eqref{eq: further_derivation}, we can calculate the influence on parameters by accumulating this term, as described in Eq.\eqref{eq: differentiate}. 
Similarly, for the influence on the loss $\mathcal{I}({z}, \mathbf{V})$, we expand it by accumulating its differences over different time steps. By applying first-order approximations of the loss and accounting for the influence of a sample $z$ on model parameters, $D^t(z)$, at different time steps $t$, we find $\mathcal{I}({z}, \mathbf{V})$ can be computed by accumulating the product of gradient of validation loss, $\nabla \mathcal{L}(\mathbf{V}, \theta^t)$, and the difference term of influence on parameters, $\hat{\mathcal{D}}^t({z})$:

\begin{algorithm*}[tp]
\caption{Calculate differential influence function for a single sample $z$.}
\label{alg}
\begin{algorithmic}[1]
{
\STATE \vspace{0.6mm} \textbf{Input:} A set of training data $\mathbf{D}$, a validation set $\mathbf{V}$, and a training sample ${z} \in \mathbf{D}$; Several checkpoints $(\theta^{t})_{t\in \{t_1, ..., t_m\}}$ during training; 
\\ 
\STATE \vspace{0.6mm} \textbf{Initialization:} $\mathcal{I}_\theta=\mathbf{0}$ and $\mathcal{I}({z}, \mathbf{V}) =0$; \vspace{0.6mm} 
\FOR{$t \in \{t_1, ..., t_m\}$} 
\STATE \vspace{0.6mm} Compute the influence difference $\mathcal{D}^t({z})$ by  Eq.\eqref{eq: further_derivation}; 
\STATE \vspace{0.6mm} $\mathcal{I}_\theta \leftarrow \mathcal{I}_\theta + \mathcal{D}^t({z})$, ~~ $\mathcal{I}({z}, \mathbf{V}) \leftarrow \mathcal{I}({z}, \mathbf{V}) + \nabla \mathcal{L}(\mathbf{V}, \theta^t)^\mathrm{T}~\mathcal{D}^t({z})$, see  Eq.\eqref{eq: theorem};  
\ENDFOR
\STATE \vspace{0.6mm} \textbf{Output:}  $\mathcal{I}_\theta$ and ~$\mathcal{I}({z}, \mathbf{V})$.
} 
\end{algorithmic}
\end{algorithm*}

\begin{proposition}
\label{tm 1}
{Let $\langle \cdot, \cdot \rangle$ denote the inner-product operation. By using the estimation for $\mathcal{D}^t({z})$ in Lemma \ref{lemma 1} and keeping the symbol convention aforementioned, the differential influence function  calculates the influence on parameters $\mathcal{I}_\theta({z})$ and the influence on loss $\mathcal{I}({z}, \mathbf{V})$ as: }
\begin{equation}
\begin{aligned}
&\mathcal{I}_\theta({z}) = \sum_{t} \hat{\mathcal{D}}^t({z}), \\
&\mathcal{I}({z}, \mathbf{V}) = \sum_t \Big\langle\nabla \mathcal{L}(\mathbf{V}, \theta^t),~~ \hat{\mathcal{D}}^t({z}) \Big\rangle. 
\label{eq: theorem} 
\end{aligned}
\end{equation} 
\end{proposition}
The proof is provided in the supplementary material. It is important to note that the naive calculation of this approximation is inefficient. This is because both the difference term in Eq.\eqref{eq: further_derivation} and the final Diff-In estimator in Eq.\eqref{eq: theorem} require summing over historical time steps, which is impractical for real-world applications. To address this, in the next subsection, we introduce practical techniques involving the use of checkpoints to significantly reduce computational costs.

\subsection{Practical Implementation Using Checkpoints}
\label{sec: Implementation Details}

\begin{figure}[http]
    \centering 
    \includegraphics[width=0.99\linewidth]{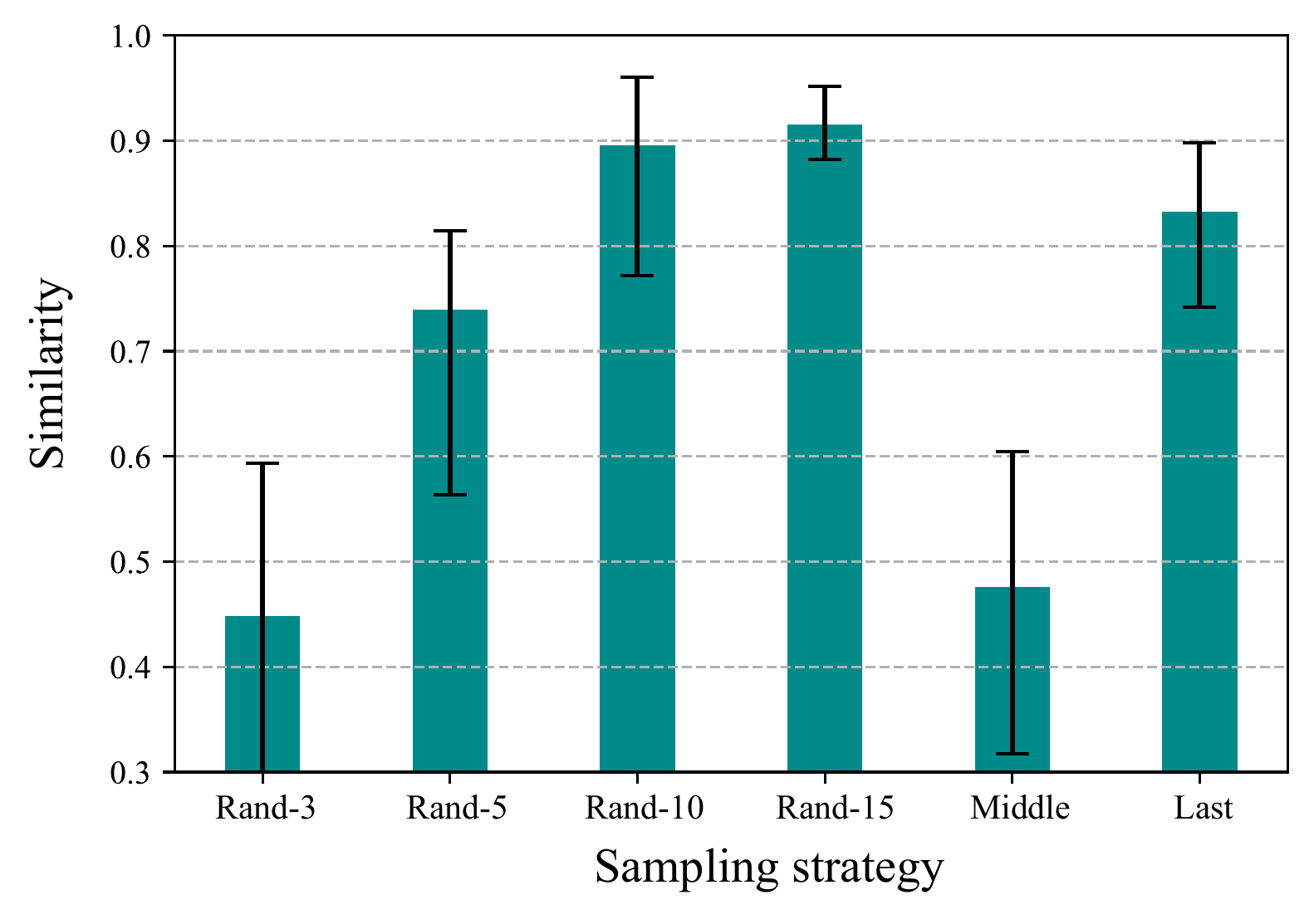}
    \vspace{-0.3cm}
    \caption{\label{fig: set k as t} Comparison of the (Cosine) similarity between the estimated difference in Eq.\eqref{eq: further_derivation} via different sampling strategies and the ground truth obtained by retraining. \textit{[Random n]} means uniformly selecting $n$ time steps. The item \textbf{\textit{[Middle]}} indicates set $k=t//2$. The item  \textbf{\textit{[last]}} indicates set $k=t$ as Eq.\eqref{eq: better further}. This experiment is done on CIFAR-10 with ResNet-18, where $t=37$.  
    }
\end{figure}

Here, we present practical methods commonly used in practice to accelerate influence estimation. 
First, for the estimator for the influence difference term in Eq.\eqref{eq: further_derivation}, we calculate it using the $t$-th checkpoint rather than all $k\leq t$ steps: 
\begin{equation}
    \hat{\mathcal{D}}^t({z}) \approx \frac{-t(\eta^t)^2}{N} \Big( H_{B^t}^t G_z^t + H_z^t ~ G_{B^t}^t \Big). 
    \label{eq: better further}
\end{equation} 
This is equivalent to when we use sampling to estimate the summation over time steps and only take the very last time step ($k=t$) as the sampled time point. We have discovered that this extremely simple operation is rather effective in the experiments. The practical performance of this strategy is quite satisfactory, see Figure \ref{fig: set k as t}.  

Second, computing Eq.\eqref{eq: theorem} requires revisiting all learning steps, which is impractical in practice. To address this, we adopt two practical strategies. Following the approach in \cite{TracIN,tan2023data}, we calculate the influence using saved intermediate checkpoints rather than all learning steps. By applying the efficient empirical rule in Eq.\eqref{eq: better further}, we have the influence calculation with checkpoints: 
\begin{equation}
\small
\begin{aligned}
    &\mathcal{I}({z}, \mathbf{V}) = \sum_{t \in \mathcal{T}_m}  \frac{-t(\eta^t)^2}{Nm} \Big\langle\nabla \mathcal{L}(\mathbf{V}, \theta^t), ( H_{B^t}^t G_z^t + H_z^t ~ G_{B^t}^t ) \Big\rangle,\\
    &\mathcal{I}_\theta({z}) =  \sum_{t \in \mathcal{T}_m} \frac{-t(\eta^t)^2}{Nm} \Big( H_{B^t}^t G_z^t + H_z^t ~ G_{B^t}^t \Big),
\end{aligned}
\end{equation} 
where $\mathcal{T}_m=\{t_1, ..., t_m\}$ is a set of uniformly selected time-steps. 
Since the calculation requires access to the batch of data $B^t$ used during training, which is often unavailable, we instead sample a random batch of data for the calculation, following \cite{TracIN}. The impact of the hyperparameter $m$ is explored in Sec. \ref{subsec: Preliminary experiments}. 
As demonstrated in the experiments in Sec.\ref{subsec: Preliminary experiments}, we found that a relatively small number of sampled time steps is sufficient to achieve strong performance. Similar strategies could be seen in \cite{TracIN, tan2023data}.

The naive calculation for Diff-In in Proposition \ref{tm 1} has a complexity of $\mathcal{O}(T^2p)$, where $T$ is the total number of training and $p$ is the number of parameters. However, since $T$ can be very large, tracing every time step would be computationally expensive and resource-intensive. 
To address this, we introduce the aforementioned time-step sampling strategy, and the overall complexity is reduced to $\mathcal{O}(mp)$. 


\vspace{0.3cm}
\textbf{Pseudo-code of Diff-In:} We summarize the calculation in Algorithm \ref{alg}. Given a selected training sample ${z}$, a validation set $\mathbf{V}$, and several time steps, we first uniformly sample $m$ time steps from the training process and select the corresponding model checkpoints $(\theta^t_\mathbf{D})_{t\in \{t_1, ..., t_m\}}$. 
Note that we can uniformly select $m$ time steps before training and save the corresponding model checkpoints during training, without saving checkpoints at all steps.  
For each time-step $t$, we calculate the influence-difference term $\mathcal{D}^t({z})$ using   Eq.\eqref{eq: differentiate}. 
Finally, after iterating over all the sampled time steps, it outputs the influence of the selected training sample ${z}$ on the model parameters and the loss on $\mathbf{V}$.

\subsection{How to Extend Diff-In to Other Optimizers?}
\label{subsec: How to Extend Diff-In to Other Optimizers?}

Thus far, the method primarily addresses models optimized with standard SGD. However, thanks to our new formulation, it can be easily extended to momentum-based gradient descent, such as SGD-M, by adjusting the parameter update equation in the derivation. 
For instance, considering the general form of gradient descent with momentum: $$\theta_{t+1} = \theta_{t} -  \eta_t \Big((1-\beta)G^t + \beta M_{t-1}\Big),$$ 
the corresponding generalized form of the estimator for $\scriptsize \hat{\mathcal{D}}^t({z})$ is: 
\begin{equation}
\small
\begin{aligned}
  &\hat{\mathcal{D}}^t({z}) \\
  &= \sum_{k<t} \alpha_k^t \Big[   \sum_{q<k} H^q \sum_{e<q} \alpha_e^q \nabla \ell(z, \theta^e)    +  \sum_{q<k}  \nabla^2 \ell(z, \theta^q)   \sum_{e<q} \alpha_e^q G^e  \Big],  
\end{aligned}
\end{equation}
where the coefficient $\alpha_k^t = \frac{1}{N}\Big(\prod_{k < a < t} \eta_a\beta_1\Big) \eta_k(1-\beta_1)$ is defined by the learning rate $\eta$ and the momentum weight $\beta$ at each step. Notably, the estimator for SGD in Lemma \ref{lemma 1} is a special case of this formulation, obtained by setting $\beta = 0$, since the key distinction between SGD and SGD-M lies in the inclusion of momentum. The detailed proof is provided in the supplementary material.

The Adam optimizer \cite{adam} uses adaptive moment estimates to adjust the learning rate for each parameter individually, resulting in faster convergence and improved performance. Given the two hyperparameters, $\beta_1$ and $\beta_2$, in Adam, we reformulate the parameter update in the form of SGD-M: 
$$\theta_{t+1} = \theta_{t} -  \eta_t^* \Big((1-\beta_1)G^t + \beta_1 M_{t-1}\Big),$$ 
where \textit{\textbf{general learning rate}} $\eta^*$ is a vector that 
$$\eta^*=\eta_t \Big/ \Big((1-\beta_1) (\sqrt{\hat{V}_t} + \epsilon)\Big),$$ 
where $\hat{V}_t = G_t^{\otimes 2}+\frac{\beta_2}{1-\beta_2} V_{t-1}$, $G_t^{\otimes 2}$ is the element-wise squaring operation on $G_t$. The vector $V$ could be easily obtained from the Adam optimizer in PyTorch \cite{pytorch}. 
Thus,  our estimator in Lemma \ref{lemma 1} by just calculating $\alpha_k^t = \frac{1}{N}\Big(\prod_{k < a < t} \eta^*_a\beta_1\Big) \eta^*_k(1-\beta_1)$ with the \textit{\textbf{general learning rate}} $\eta^*$.

\section{Approximation Error}
\label{sec:theoretical}

We analyze the approximation error of the proposed influence estimator in Proposition \ref{tm 1}, where 
$\mathcal{I}(z)$ is the estimated influence and $\mathcal{I}^\text{Exact}(z)$ is the exact influence calculated by the vanilla leave-one-out retraining:

\begin{proposition}
\label{tm 2}
{Supposing the loss has $\ell$-Lipschitz continuous gradient and the gradient norm of the parameter is upper-bounded by $g$, and assuming the learning rate $\eta \leq 1$ and the momentum weight $\beta \leq 1$ (if available), the error between the approximated $\mathcal{I}(z)$ and the exact $\mathcal{I}^\text{Exact}(z)$ is bounded by:  }
\begin{equation}
|\mathcal{I}(z) - \mathcal{I}^\text{Exact}(z)| \leq 2 T^2\ell ( T + 1) C + \frac{T^2}{N}g, 
 \label{eq: theorem 2}
\end{equation}
{where $|\cdot|$ is the norm, and $T$ is the maximum iteration, $N$ is the number of training samples, 
$C$ represents the farthest distance the neural network parameters move away from their initial state during training when any subset $\mathbf{D}_s \subset \mathbf{D}$ is used as the training set. The detailed derivation is provided in the supplementary material.}
\end{proposition}

\subsection{Discussion of the polynomial error bound} The bound proposed in Proposition \ref{tm 2} shows that the upper bound of the approximation error of Diff-In grows with the increase of the training times $T$. 
The reason that increasing the training steps leads to larger errors is that the optimized parameters diverge further from the initialized parameters as training progresses. This divergence makes accurate estimation more challenging and contributes to the accumulation of approximation errors. 
It is also worth noting that some methods, such as those in SGD-Inf \cite{hara2019data} and data-value-embedding influence (DVE-INF) \cite{wang2025capturingDVE}, exhibit a faster increase in error as $T$ grows. In contrast, our approach, with its polynomial error bound, demonstrates a significantly smaller growth rate compared to other methods, such as the exponential growth observed with SGD-Inf \cite{hara2019data}. This indicates that our method effectively mitigates error accumulation over time, even as $T$ increases. It is important to emphasize that this bound reflects a worst-case scenario for error. In practice, Diff-In performs robustly even with larger $T$ values. As training progresses and the model approaches convergence, parameter changes become minimal, meaning larger $T$ typically has a negligible impact on errors.  Here, we show the performance change of the coreset selected by Diff-In vs the change of $T$ on ImageNet, see Table \ref{tb:duration ablation}. When $T=50$ (the default setting in this paper), the top-1 accuracy of the model trained on the coreset is 61.7. When $T=100$ and $T=200$, the performance is 61.4 and 62.4, respectively. Notably, increasing $T$ does not degrade Diff-In's practical performance.

\begin{table}[ht]
\centering
\caption{\label{tb:duration ablation}
The performance change of the coreset selected by Diff-In v.s. the change of training duration $T$ (epochs). This experiment is conducted on ImageNet \cite{imagenet} with ResNet-50 \cite{resnet}. 
}
\begin{tabular}{@{}ccc@{}}
\toprule
\textbf{Training duration $T$ (epochs)} & \textbf{Top-1 Accuracy (\%)} \\ 
\midrule
50  & 61.7 \\ 
100 & 61.4 \\ 
200 & 62.4 \\ 
\bottomrule
\end{tabular}
\end{table}

\subsection{Discussion of the smoothness assumption} Lipschitz continuity and gradient norms are commonly used to characterize the smoothness of a neural network's loss landscape. Modern deep learning models incorporate techniques such as normalization layers (e.g., batch normalization and layer normalization) and shortcut connections to enhance smoothness and continuity \cite{how_bn,how_resnet}, facilitating optimization. These techniques make the assumption of smoothness generally valid in practice. This stands in contrast to many studies \cite{IF,Studying_large_language} that rely heavily on the convexity of neural networks. Our assumption, by comparison, is much easier to satisfy in real-world scenarios. 
If the conditions for $\ell$-Lipschitz continuous gradients are not met, the gradient norms $g$, the values of $g$, and $\ell$ can become very large. In such cases, the error bound derived from these parameters may lose its practical relevance. However, this does not necessarily imply that the algorithm will fail in practice.

\section{Experiments} \label{sec:experiment}

We extensively evaluate our method on three major data-centric tasks: dataset cleaning {(Sec.\hspace{0.4mm}\ref{sec: experiment cleaning})}, data deletion {(Sec.\hspace{0.4mm}\ref{subsec:data_deletion})}, and coreset selection {(Sec.\hspace{0.4mm}\ref{subsec:exp_coreset_selection})}. Then, we conduct ablation studies and analysis on Diff-In (Sec.\hspace{0.4mm}\ref{subsec: Preliminary experiments}). Our experiments were mainly conducted with Pytorch \cite{pytorch} on two servers, each with 8 Tesla-H200 GPUs.

\subsection{Datasets, Setups, and Baselines}
\label{sec: Datasets and General Settings}

\vspace{0.13cm}
\textbf{Datasets:} We introduce the datasets utilized in this study. For the three tasks mentioned above, we conducted experiments based on conventional classification tasks, using the following datasets: (1). \textbf{CIFAR-100} \cite{CIFAR} contains 50,000 training images and 10,000 test images across 10 distinct classes. 
(2). \textbf{Tiny-ImageNet} \cite{tiny} comprises 100,000 images belonging to 200 classes. 
(3). \textbf{ImageNet-1K} \cite{imagenet} encompasses 1,000 classes and includes over 1 million training images.

Subsequently, we carried out experiments in the context of large models. For data cleaning and data deletion, we used the \textbf{GSM8K} (Grade School Math 8K) dataset. For coreset selection, we opted for a larger and more challenging vision-language dataset: \textbf{CC12M} \cite{CC12M}, which includes 12 million image-text pairs sourced from the Internet, facilitating CLIP-like vision-language pre-training \cite{CLIP}.

\vspace{0.13cm}
\textbf{General setups:} Note that when approximating Diff-In, we need to compute the information over the entire dataset, \textit{e.g.}, the gradient $G=\nabla \mathcal{L}(\mathcal{D}, \theta)$ over the whole dataset. 
In practice, we use a random batch as an efficient proxy for the entire dataset. This random batch has a size of $2048$ for experiments on ImageNet-1K \cite{imagenet} while $512$ for others. Similar proxy schemes were also adopted in previous works \cite{IF,tan2023data,TracIN,OPT}. All the results are averaged over 5 independent runs. More details on the implementation and experiments are reported in the supplementary material.

\vspace{0.13cm}
\textbf{Baselines:} We selected several typical influence estimators, including Koh's method (referred to as IF) \cite{IF} and its subsequent works, such as DataInf \cite{kwon2023datainf}, EK-FAC \cite{Studying_large_language}, and GEX \cite{GEX}. Then, we select the famous TracIn \cite{TracIN} and TRAK \cite{Attribution_acale} as baselines. Due to the extreme time consumption of the SGD-INF computation process (takes more than 100 seconds for each 64×64‑pixel small image), we were unable to run it across numerous experiments, especially in larger-scale scenarios. Therefore, we opted for the more recent DVE-INF \cite{wang2025capturingDVE}, which employs strategies like random projection to enhance speed. 
It is worth noting that in the data deletion experiments, all methods mentioned above, except for IF \cite{IF}, could not be applied, as they do not provide estimates of the data's influence on parameters. For different tasks, we will also select some baselines that are not based on influence for comparison.

\subsection{Data Cleaning} 
\label{sec: experiment cleaning}

\begin{table*}[tp]
\caption{\label{tb: data cleaning 1} 
Experimental results (Precision) of data cleaning on classification datasets. The best results are bolded. 
Here, Diff-In estimates the sample's influence on the validation loss (validation influence) and its influence on its own loss (self-influence). 
}
\setlength{\tabcolsep}{3.1pt}
\centering
\resizebox{0.99948\linewidth}{!}{  
\begin{tabular}{lcccccccccccccccccc}
    \toprule
    \textbf{Dataset ($\rightarrow$)} & \multicolumn{6}{c}{\textbf{CIFAR-100} \cite{CIFAR}} & \multicolumn{6}{c}{\textbf{Tiny-ImageNet} \cite{tiny}} & \multicolumn{6}{c}{\textbf{ImageNet-1K} \cite{imagenet}} \\
    \cmidrule(lr){2-7} \cmidrule(lr){8-13} \cmidrule(lr){14-19}
    \textbf{Selection Rate ($\rightarrow$)} & \textbf{20\%} & \textbf{30\%} & \textbf{40\%} & \textbf{50\%} & \textbf{60\%} & \textbf{70\%} & \textbf{20\%} & \textbf{30\%} & \textbf{40\%} & \textbf{50\%} & \textbf{60\%} & \textbf{70\%} & \textbf{20\%} & \textbf{30\%} & \textbf{40\%} & \textbf{50\%} & \textbf{60\%} & \textbf{70\%}\\
    \midrule
    \textbf{Random} & 20.0 & 30.0 & 40.0 & 50.0 & 60.0 & 70.0
   & 20.0 & 30.0 & 40.0 & 50.0 & 60.0  & 70.0
    & 20.0 & 30.0 & 40.0 & 50.0 & 60.0 & 70.0  \\
    \textbf{Loss value} \cite{choromanska2015loss} &28.7 &49.5 &53.2 &69.1 &72.8 &74.6
    &21.2 &31.7 &42.1 &52.2 &62.6 &65.3
    & 19.5 & 32.4 & 41.7 & 52.8 & 62.1 & 72.9
    \\
    \midrule
    \textbf{IF} \cite{IF} &42.2 &45.0 &58.4 &65.3 &73.8 &78.3
     &21.1 &31.6 &42.0 &52.2 &62.3 &69.4
     & 19.1 & 36.2 & 40.6 & 52.8 & 62.7 & 70.1\\
     \textbf{DataInf} 
    \cite{kwon2023datainf}  &45.2 &49.7 &60.8 &69.5 &75.3 &80.4
     &26.9 &38.6 &45.5 &58.6 &64.8 &70.5 
     & 21.7 & 38.3 & 41.5 & 55.6 & 63.4 & 72.8 \\
\textbf{EK-FAC} \cite{Studying_large_language}
&42.8 &50.3 &60.9 &72.0 &75.6 &80.2 
     &25.4 &37.2 &43.6 &57.9 &63.7 & 70.6 
      & 17.2 & 34.7 & 39.2 & 51.3 & 60.8 & 68.3
    \\
     \textbf{GEX} \cite{GEX} &{80.0} &84.3 &87.5 &{89.0} &92.0 &94.1
&{72.7}  &77.2  &81.9  &87.3  &90.1 &92.2
& 23.1 & 34.0 & 44.8 & 54.5 & 63.6 & 69.7
  \\
     \textbf{TracIN} \cite{TracIN} 
&{79.5} &{88.7} &{93.9} &{96.5} &{97.3} &98.1   
&76.3 &{91.0} &{95.9} &{97.8} &{98.4}    &99.3
& 46.4 & 52.2 & 59.1 & 67.2 & 70.2 & 78.0
 \\
    \textbf{DVE-INF} \cite{wang2025capturingDVE}  
    &81.3 &92.1 &93.4 &95.6 &95.8 &96.1
     &82.0 &92.4 &97.1 &98.8 &99.5 &99.9 
      & 44.8 & 50.7 & 58.4 & 66.0 & 71.6 & 79.5 \\
     \midrule
     \textbf{Diff-In (Validation Influence)} 
     &{84.3} &{92.8} &{95.1} &{96.4} &{97.8} &{98.6} 
     &{87.9} &{94.7} &{97.9} &{98.4} &{98.9} &{99.0} 
     & 47.2 & 54.1 & 60.4 & 67.8 & 71.0 & 78.6\\
     \textbf{Diff-In (Self-Influence)} 
     &\textbf{91.4} &\textbf{93.9} &\textbf{97.5} &\textbf{98.2} &\textbf{99.1} &\textbf{99.4} 
     &\textbf{88.2} &\textbf{98.0} &\textbf{98.6} &\textbf{99.0} &\textbf{99.8} &\textbf{100.0} 
     & \textbf{53.6} & \textbf{58.8} & \textbf{64.1} & \textbf{69.6} & \textbf{74.2} & \textbf{81.5}\\
\bottomrule
\end{tabular}
} 
\end{table*}

Data cleaning \cite{imagenetclean, Shapley_noise_clean_zou} is a crucial step in the data pre-processing pipeline aimed at improving the quality and reliability of dataset labels. In our experiments, we define data cleaning as the process of identifying mislabeled samples within the dataset. Detailed experimental settings can be found in the supplementary material. Here, we introduce two kinds of influence metrics for assessing label quality, namely self-influence and validation influence:  

\begin{enumerate}
    \item The self-influence metric examines a sample's influence on its own loss \cite{TracIN}: 
\begin{equation}
\label{eq: self influence original}
    \mathcal{I}(z, z) = \mathcal{L}(z, \theta^{*}_{-z}) - \mathcal{L}(z, \theta^{*}).\\
\end{equation}
According to the definition of a single data point's influence on the loss function (Eq.\eqref{eq: original influence on params}), it is just setting the validation set as the sample $z$ itself. 
Self-influence plays a key role in identifying mislabeled points \cite{TracIN, IF}. Outliers or mislabeled data can significantly impact their own loss, or self-influence \cite{TracIN}, because the model may not receive sufficient information from other samples to handle these outliers effectively, leading to higher self-influence values for such samples.  
    \item The validation influence measures the influence of a sample on the validation loss: 
\begin{equation}
\label{eq: self influence original}
    \mathcal{I}(z, V) = \mathcal{L}(V, \theta^{*}_{-z}) - \mathcal{L}(V, \theta^{*}).\\
\end{equation} 
Theoretically, incorrect labeling can lead to an increase in validation loss, indicating that the validation influence of the corresponding sample should be negative. 
\end{enumerate} 

We selected several well-known influence estimators as baselines, including IF \cite{IF}, GEX \cite{GEX}, and DataInf \cite{kwon2023datainf}, which lack training-dynamic awareness, as well as TracIn \cite{TracIN} and DVE-INF \cite{wang2025capturingDVE}, which do incorporate training-dynamic awareness. Additionally, we chose classic loss value \cite{choromanska2015loss} methods from data cleaning as an extra baseline. 

\vspace{0.3cm}
\subsubsection{Experiments on classification datasets}  
We perform experiments on several image classification datasets, namely CIFAR-100 \cite{CIFAR}, Tiny-ImageNet \cite{tiny}, and ImageNet-1k \cite{imagenet}. To create a noisy dataset, we uniformly select 20\% of the data and perturb their corresponding labels by following \cite{TracIN,GEX}.  Then, we train a ResNet model \cite{resnet} using the contaminated data (ResNet-18 for CIFAR and Tiny-ImageNet, ResNet-50 for ImageNet-1K). We use the parameters of this model to calculate the influence of the samples. Additionally, other baseline influence estimators, such as IF \cite{IF}, TracIn \cite{TracIN}, GEX \cite{GEX}, DataInf \cite{kwon2023datainf}, and DVE-INF \cite{wang2025capturingDVE}, are also employed to compute the self-influence metric. The experimental results are presented in Table \ref{tb: data cleaning 1}, where we report the \textbf{Precision} metric under different selection ratio settings, $\text{Precision} = \frac{N_\text{found}}{N_\text{all}}$,  
where $N_\text{found}$ indicates the number of noise samples found by algorithms and $N_\text{all}$ is the number of all noise samples.

Diff-In outperforms the compared methods in all scenarios. For instance, considering the challenging Tiny-ImageNet dataset,  many baselines, including the IF \cite{IF}, DataInf \cite{kwon2023datainf}, EK-FAC \cite{Studying_large_language}, and loss-based method \cite{choromanska2015loss}, only achieve marginal improvements over random selection. When employing Diff-In to select the top 20\% samples, it is possible to identify over 88\% of the erroneous samples, exceeding others like TracIN by approximately 12.0\%. 
The experimental results on ImageNet-1K \cite{imagenet} show that TracIn \cite{TracIN}, DVE-INF \cite{wang2025capturingDVE}, and our Diff-In, as algorithms that are sensitive to training dynamics, significantly outperform the remaining methods that lack this sensitivity. Notably, our approach consistently exceeds TracIn and DVE-INF by approximately 2.0 to 8.1 percentage points across various selection rates. Moreover, we also implement the Diff-In for the validation influence for data cleaning. A slight drop in performance was observed, suggesting that self-influence may be a more effective method for identifying incorrect or outlier data. Additionally, self-influence is computationally more efficient, as it requires only the sample itself to calculate influence, rather than relying on a separate validation dataset.

\begin{table}[tp]
\caption{\label{tb: data cleaning large models} 
Experimental results (Precision) of data cleaning on GSM8K. The best results are bolded.  
Here, Diff-In estimates the sample's influence on the validation loss (validation influence) and its influence on its own loss (self-influence). }
\setlength{\tabcolsep}{3.1pt}
\centering
\resizebox{0.99948\linewidth}{!}{  
\begin{tabular}{lcccccccccccccccccc}
    \toprule
    \textbf{Dataset ($\rightarrow$)}  & \multicolumn{6}{c}{\textbf{GSM8K} \cite{cobbe2021gsm8k}} \\
    \cmidrule(lr){2-7}  
    \textbf{Selection Rate ($\rightarrow$)}  & \textbf{20\%} & \textbf{30\%} & \textbf{40\%} & \textbf{50\%} & \textbf{60\%} & \textbf{70\%}\\
    \midrule
    \textbf{Random} 
    & 20.0 & 30.0 & 40.0 & 50.0 & 60.0 & 70.0  \\
    \textbf{Loss value} \cite{choromanska2015loss} 
    &28.4 &39.3 &51.5 &57.4 &63.7 &69.4
    \\
    \midrule
    \textbf{IF} \cite{IF} 
     &67.2 &71.4 &79.6 &88.5 &94.7 &97.7\\
     \textbf{DataInf} 
    \cite{kwon2023datainf}  
     &68.5 &73.4 &84.2 &90.1 &96.2  &98.5 \\
    \textbf{EK-FAC} \cite{Studying_large_language} 
      &67.1 &73.1 &82.2 &90.4 &94.9 &98.0 
    \\
     \textbf{GEX} \cite{GEX} 
&77.2 &{87.1} &{91.6} &{95.3} &{99.9} &100.0
  \\
     \textbf{TracIN} \cite{TracIN}  
&78.4 &{84.5} &{92.2} &{99.2} &{99.2} &100.0
 \\
    \textbf{DVE-INF} \cite{wang2025capturingDVE}   
      &80.1 &86.8 &93.4 &98.6 &99.1 &99.5 \\
     \midrule
     \textbf{Diff-In (Validation Influence)} &{85.8} &{90.4} &{98.6} &{99.2} &{99.8} &{99.8}\\
     \textbf{Diff-In (Self-Influence)} &\textbf{86.1} &\textbf{92.2} &\textbf{99.3} &\textbf{99.9} &\textbf{99.9} &\textbf{100.0}\\
\bottomrule
\end{tabular}
}
\vspace{-10pt}
\end{table}

\vspace{0.3cm}
\subsubsection{Experiments on large models}  Furthermore, we present the results of data cleaning experiments conducted on the SFT dataset GSM8K \cite{cobbe2021gsm8k} for various approaches utilizing large language models \cite{dubey2024llama3herdmodels}. Consistent with the classification experiments, we selected the same baselines and maintained a data noise perturbation ratio of 20\%. Here we provide an example problem from GSM8K:

\textit{\textcolor{teal}{Question: Beth bakes 4.2 dozen batches of cookies in a week. If these cookies are shared equally among 16 people, how many cookies does each person consume?}} 

\textit{\textcolor{teal}{Answer: 6}} 

\noindent where we will randomly change the answer in the above text to other numbers (for example, change the answer 6 to 9). In our experiments, we LoRA-finetuned a Llama 3.1 8B model \cite{dubey2024llama3herdmodels} on the contaminated GSM8K dataset \cite{cobbe2021gsm8k} and subsequently used the LoRA parameters to calculate influence, following the methodologies outlined in \cite{xia2024less,kwon2023datainf}. We computed the influence values for each sample, including both self-influence and validation influence. Finally, we observe how each influence estimator serves as an indicator for assessing the performance of noisy data. The experimental results, presented in Table \ref{tb: data cleaning large models}, focus on the \textbf{Precision} metric across various selection ratio settings.

The experimental results presented in Table \ref{tb: data cleaning large models} demonstrate the precision of various data cleaning methods on the GSM8K dataset across different selection rates. The table highlights the performance of Diff-In (both Validation Influence and Self-Influence variants) compared to other state-of-the-art approaches. Diff-In (Validation Influence) consistently achieves high precision scores, ranging from 85.8\% at a 20\% selection rate to 99.8\% at a 70\% selection rate. Diff-In (Self-Influence) performs even better, achieving near-perfect precision scores, with values of 86.1\% at 20\% and achieving 100\% after the 50\% selection rate. This is consistent with the conclusions from the classification experiments, suggesting that self-influence may be a more effective method for identifying incorrect or outlier data. Training-dynamic aware methods, such as TracIN \cite{TracIN} and DVE-INF \cite{wang2025capturingDVE}, demonstrate strong performance compared to traditional loss value methods. In contrast, non-training-dynamic aware approaches like IF \cite{IF}, DataInf \cite{kwon2023datainf}, and EK-FAC \cite{Studying_large_language} perform less effectively. Our training-dynamic aware approach consistently outperforms other methods, including TracIN \cite{TracIN} and DVE-INF \cite{wang2025capturingDVE}, across various settings. Notably, at lower selection rates (20\% and 30\%), our method surpasses both by 4.0 to 6.0 percentage points.

\begin{table*}[http]
\newcommand{\tabincell}[2]{\begin{tabular}{@{}#1@{}}#2\end{tabular}}
\setlength\tabcolsep{3.1pt}
\centering 
\caption{\label{tb:data_deletion}  
The experimental results of data deletion, where the performance metric is Accuracy@1, where the column "Noise" contains results on the noise set, while the column "Oracle" contains results retraining on the cleaned set (all noise samples are removed).  
Here, Diff-In estimates the sample's influence on the parameters. 
} 
\resizebox{0.789999\linewidth}{!}{
\begin{tabular}{lcccccccc}
    \toprule
    ~ & \multicolumn{3}{c}{\textbf{Image Classification}} & \textbf{Large Models}\\ 
        \cmidrule(lr){2-4} \cmidrule(lr){5-5} 
\textbf{Dataset ($\rightarrow$)}  &$~~$\textbf{CIFAR-100} \cite{CIFAR}  &$~~$\textbf{Tiny-ImageNet} \cite{tiny} &$~~$\textbf{ImageNet-1K} \cite{imagenet} &$~~$\textbf{GSM8K} \cite{cobbe2021gsm8k}  \\
\midrule[0.8pt]
\textbf{Noise (training with noise)} &70.7  &39.7 &61.3 &81.7\\
\textbf{Oracle (training without noise)} & \textbf{74.1}     &\textbf{45.7} &\textbf{72.9} &\textbf{92.3}\\
\midrule[0.1pt]
\textbf{IF} \cite{IF} &$70.4$ &$38.2$ &$64.3$ &$82.5$\\
\textbf{MC-IF} \cite{unlearning_iclr} &$71.1 $  &$40.3 $ &$65.1$ &$84.1 $ \\
\textbf{SSSE} \cite{SSSE_unlearning} &$71.9 $  &$41.5 $ &$67.2$ &$85.7 $ \\
\midrule[0.1pt]
\textbf{Diff-In} (ours) &$\textbf{73.6} $      &$\textbf{42.9}$  &$\textbf{68.3}$ &$\textbf{89.9} $ \\ 
\bottomrule[0.8pt]
\end{tabular}  
}
\end{table*}

\subsection{Data Deletion: Removing the influence of noisy data without retraining} 
\label{subsec:data_deletion}

Data deletion, also known as machine unlearning \cite{unlearning_iclr}, is the task of eliminating the impact of certain training data items from a learned model. It is noteworthy that some methods in this field rely on retraining, while others do not. Diff-In falls into the latter category. To ensure fairness, the baselines discussed here exclusively consist of methods that do not depend on retraining. 
Considering the physical meaning of the influence $\mathcal{I}_\theta(z) = \theta^*_{-z} -\theta^*$, we can approximately obtain the leaving-one-retraining parameters by $\theta^*_{-z} = \mathcal{I}_\theta(z) + \theta^*$ after calculating the approximation $\mathcal{I}_\theta(z)$. Hence, $\theta^*_{-z}$ is the approximated parameter after performing data deletion. For deleting a set of samples $\mathbf{Z}$ from the model, according to the additivity assumption of influence functions \cite{IF,group_rffect_analysis,OPT,tan2023data}, we can add up their influence as a whole, specifically, $\theta^*_{-\mathbf{Z}} = \sum_{z \in \mathbf{Z}} \mathcal{I}_\theta(z) + \theta^*$. Note that many methods \cite{TracIN,GEX,Mirrored,kwon2023datainf} do not provide the influence on parameters, $\mathcal{I}(\theta)$, and therefore cannot be applied in this experiment. 

It is important to note that many methods, such as TracIn \cite{TracIN} and GEX \cite{GEX}, cannot be applied to this task because they are unable to estimate the influence of samples on parameters. Additionally, methods like DataInf and EK-FAC, which are acceleration approaches based on IF, do not provide more accurate estimations. Therefore, we did not include them as baselines in this context. Thus, we selected IF \cite{IF} along with two other data deletion methods (MC-IF \cite{unlearning_iclr} and SSSE \cite{SSSE_unlearning}) as the baseline for comparison.

\vspace{0.3cm} 
\subsubsection{Experiments on classification datasets}

Herein, we employ influence estimators to mitigate the impact of noisy samples on the learned model. For introducing noise, we follow the settings outlined in Sec.\ref{sec: experiment cleaning}, specifically introducing 20\% label noise to the CIFAR-100 \cite{CIFAR}, Tiny-ImageNet \cite{tiny} image, and ImageNet-1K \cite{imagenet}, three classification datasets. Then, we train a ResNet model \cite{resnet} using the contaminated data (ResNet-18 for CIFAR and Tiny-ImageNet, ResNet-50 for ImageNet-1K). Finally, we will employ various methods to minimize the impact of noisy data on the performance of the already trained model, all without retraining. 
To execute data deletion for noise samples, we compute the influence on parameters (denoted by $\mathcal{I}_\theta(\mathbf{D}_n)$) of the data to be deleted, where $\mathbf{D}_n$ signifies the subset encompassing all the noisy data. 
Note that this experiment is different from the Data Cleaning shown above. Data Cleaning aims to find mislabeled data, while the purpose of Data deletion here is: given that the model has already been trained on a dataset, use some methods to eliminate the influence of particular data samples without retraining.

Apart from the simplistic retraining method (referred to as the Oracle), we also engage in a comparison with the Influence Function estimator (abbreviated as IF) put forward by Koh and Liang \cite{IF} and the MCMC-IF \cite{unlearning_iclr}, SSSE  \cite{SSSE_unlearning} as baseline approaches. The experimental results are presented in Table \ref{tb:data_deletion}. 
Diff-In achieves impressive accuracies of 72.6\%, 42.9\%, and 68.3\% on CIFAR-100, Tiny-ImageNet, and ImageNet-1K, respectively, outperforming the second-best method, SSSE \cite{SSSE_unlearning}, by 1.7\%, 1.4\%, and 1.1\% on these three datasets. Notably, the performance gap compared to the Oracle brute-force retraining on CIFAR-100 is only about 0.5\%, underscoring its advantages over other approaches.

\vspace{0.3cm} 
\subsubsection{Experiments on large models} 

Next, we conducted data deletion experiments on the GSM8K dataset \cite{cobbe2021gsm8k} using the Llama 3.1 8B model \cite{dubey2024llama3herdmodels}. We began by randomly contaminating 20\% of the labels in GSM8K, employing a pollution strategy consistent with our data cleaning approach. After this, we LoRA-finetuned the Llama 3.1 8B model \cite{dubey2024llama3herdmodels} on the contaminated dataset and used the LoRA parameters to calculate influence before executing the data deletion process.

The relevant experimental results are presented in Table \ref{tb:data_deletion}. Consistent with the findings from classification tasks, Diff-In demonstrates significantly better data deletion performance on large models, achieving 89.9\% accuracy, which is only 2.4\% lower than the 92.3\% performance of the oracle brute-force retraining on the clean GSM8K. It significantly outperforms other methods, such as SSSE \cite{SSSE_unlearning} by 4.2\% and MC-IF \cite{unlearning_iclr} by 5.8\%. This highlights Diff-In’s exceptional ability to estimate the influence of samples on model training.


\begin{table*}[tp]
\caption{\label{tb: coreset} 
Experimental results (Accuracy@1) of coreset selection on three image classification datasets.  
The best results are bolded for baselines and ours, respectively. 
Here, Diff-In estimates the sample's influence on the training loss \cite{tan2023data}. 
}
\setlength{\tabcolsep}{3.1pt}
\centering
\resizebox{0.85188\linewidth}{!}{  
\begin{tabular}{lcccccccccccccccccc}
    \toprule
    \textbf{Dataset ($\rightarrow$)} & \multicolumn{6}{c}{\textbf{CIFAR-100}} & \multicolumn{6}{c}{\textbf{Tiny-ImageNet}} & \multicolumn{6}{c}{\textbf{ImageNet-1K}} \\
    \cmidrule(lr){2-7} \cmidrule(lr){8-13} \cmidrule(lr){14-19}
    \textbf{Selection Rate ($\rightarrow$)}  & \textbf{20\%} & \textbf{30\%} & \textbf{50\%} & \textbf{70\%} & \textbf{80\%} & \textbf{100\%} & \textbf{20\%} & \textbf{30\%} & \textbf{50\%} & \textbf{70\%} & \textbf{80\%} & \textbf{100\%} & \textbf{20\%} & \textbf{30\%} & \textbf{50\%} & \textbf{70\%} & \textbf{80\%} & \textbf{100\%}\\
    \midrule
    \textbf{Random}  
     & 50.2 & 53.6 & 64.3 & 71.0 & 74.1 & 78.1  
    & 24.0 & 29.7 & 34.4 & 40.9 & 45.7   & 49.3    
     & 61.6 & 65.9 & 67.7 & 70.3 & 72.9 & 76.4 \\
    \midrule
     \textbf{SSP}  \cite{SSP}
    & 44.4 & 54.6 & 62.9 & 70.7 & 75.2  
    &- & 20.8 & 27.6 & 32.5 & 39.6 & 44.9   
    &- & 31.2 & 51.4 & 60.3 & 69.8 & \textbf{75.5} &-   \\
    \textbf{Moderate}     \cite{Moderate}
    & {51.8} & {57.7} & {64.9}  & \textbf{71.8}  & 74.2    
    & -  & {25.2} & \textbf{30.5}  & {34.8}  & \textbf{41.4}  & 46.0   
    &-  & 61.1 & {67.8} & {70.0} & {73.0} & 75.3 &-  \\
    \midrule 
    \textbf{IF} \cite{IF}   & 26.4 & 36.1 & 47.1 & 51.2 & 63.5  
    &- & 14.3 & 19.1 & 24.5 & 31.2 & 37.9   
    &- & 25.6 & 30.5 & 49.9 & 56.7 & 68.1 &-   \\  
    \textbf{DataInf} \cite{kwon2023datainf}  & 26.9 & 35.8 & 48.2 & 52.5 & 65.5  
  &- & 16.1 & 21.2 & 27.7 & 33.1 & 42.4   
  &- & 27.7 & 32.3 & 52.6 & 58.2 & 67.0  &-  \\  
    \textbf{EK-FAC} \cite{Studying_large_language}  &27.8 &37.1 &47.5 &52.4 &66.2 &- 
     &17.5 &20.2 &25.6 &32.7 &38.0 &- 
      &25.9 &31.9 &51.6 &58.3 &69.1 &- \\
    \textbf{GEX} \cite{GEX} & 44.8 & 50.0 & {57.5} & {62.0} & {69.9} 
    &- & {20.5} & {23.0} & {29.4}  & 36.6  & {41.6}
    &- & {55.3} & {58.7} & {65.7} & 69.8 & {70.1} &-   \\
    \textbf{TRAK} \cite{Attribution_acale} &49.5 &52.9 &59.6 &70.4 & 73.9 &-
     &26.2 &29.3 &35.8 &39.1 &45.0 &- 
      &58.4 &64.5 &68.6 &70.1 &72.9 &- \\
    \textbf{TracIn} \cite{TracIN} & 50.4 & 58.4 & {63.9} & {70.4} & {72.3} 
    &- & {24.2} & {28.8} & {35.3}  & 40.1  & {45.8}
    &- & {61.3} & {64.4} & \textbf{70.3} & 71.4 & {73.6} &-  \\  
    \textbf{DVE-INF}\cite{wang2025capturingDVE}  & 51.2 & 57.0 & 63.6 & 70.2 & 73.3 & - 
    &25.1 &29.1 &34.7 & 40.5 & 44.9 & -
    &60.5 &65.2 &67.7 &72.9 & 73.5 & - \\
    \midrule 
    \textbf{Diff-In}    
    & \textbf{52.0} & \textbf{59.4} & \textbf{65.7} & {71.5} & \textbf{75.3} 
    &- & \textbf{26.6} & \textbf{30.5} & \textbf{36.0}  & 40.9  & \textbf{46.4}
    &- & \textbf{61.7} & \textbf{68.3} & {69.8} & \textbf{73.6} & \textbf{75.5} &-   \\
\bottomrule
\end{tabular}
}
\end{table*}

\begin{table*}[tp]
\caption{\label{tb: coreset clip} 
Experimental results of coreset selection on vision-language pertaining. 
Regarding Diff-In, we contemplate two time-step configurations: uniformly selecting times (designated as Diff-In, the normal setting) and selecting only the final time step (designated as Diff-In-F, the efficient setting). 
The top-2 highest results are bolded. 
Here, Diff-In estimates the sample's influence on the training loss \cite{tan2023data}. }
\setlength{\tabcolsep}{3.1pt}
\centering
\resizebox{0.9988\linewidth}{!}{  
\begin{tabular}{lcccccccccccccccccccccccc}
    \toprule
    \textbf{Dataset ($\rightarrow$)} & \multicolumn{5}{c}{\textbf{Zero-shot Classification} (Acc@1)} & \multicolumn{5}{c}{\textbf{I2T Retrieval} (Recall@1)} & \multicolumn{5}{c}{\textbf{T2I Retrieval} (Recall@1)} & \multicolumn{5}{c}{\textbf{Linear Prob} (Acc@1)} \\
    \cmidrule(lr){2-6} \cmidrule(lr){7-11} \cmidrule(lr){12-16} \cmidrule(lr){17-21}
    \textbf{Selection Rate ($\rightarrow$)} & \textbf{10\%} & \textbf{20\%} & \textbf{40\%} & \textbf{60\%} & \textbf{100\%} & \textbf{10\%} & \textbf{20\%} & \textbf{40\%} & \textbf{60\%} & \textbf{100\%}& \textbf{10\%} & \textbf{20\%} & \textbf{40\%} & \textbf{60\%} & \textbf{100\%} & \textbf{10\%} & \textbf{20\%} & \textbf{40\%} & \textbf{60\%} & \textbf{100\%}\\
    \midrule 
    \textbf{Random} 
    & 19.7 & 21.3 & 25.3 & 29.0  & 37.2
    & 29.9 & 32.1 & 38.1 & 42.3  & 51.7
    & 21.2  &23.1  & 25.4 & 28.6   & 44.5
    & 41.6 & 45.4 & 51.3 & 59.1   &67.5 \\
    \midrule[0.8pt]
    \textbf{SSP} \cite{SSP}
    & 18.0 & 20.4 & 23.2 &28.4  &-
    & 24.5 & 31.8 & 35.0 & 39.4  &-
    & 21.1 & 22.5 & 24.8 & 29.2  &-
    & 40.8 & 44.6 & 52.0 & 58.4  &-\\
    \textbf{CLIP-score} \cite{CLIP}
    & 24.1 & 27.4 & 32.4 & 34.1  &-
    & 33.9 & 39.1 & 45.2 & 47.1 &-
    &\textbf{23.7}  &28.4  & 36.4 & 40.3  &-
    & 47.4 & 52.9 & 59.3 & 62.1   &-\\    
    \textbf{Moderate}  \cite{Moderate}
    & 24.7 & 27.7 & 31.7 & 32.0  &-
    & 33.4 & 37.5 & 41.0 & 43.7 &-
    & 22.7  &25.8  & 34.2 & 38.9  &-
    & 47.9 & 51.4 & 56.5 & 60.2   &-\\
    \midrule[0.8pt] 
    \textbf{IF}   \cite{IF}
    & 22.3 & 24.8 & 29.0 & 30.4  &-
    & 30.4 & 37.3 & 42.5 & 45.9 &-
    & 20.8  &25.2  & 30.4 & 35.7  &-
    & 43.3 & 47.1 & 54.7 & 60.7   &-\\  
    \textbf{DataInf}   \cite{kwon2023datainf}
    & 22.5 & 25.4 & 29.8 & 31.3  &-
    & 30.1 & 37.2 & 43.3 & 46.7 &-
    & 21.5  &25.9  & 31.9 & 35.6  &-
    & 43.8 & 48.3 & 55.1 & 59.4   &-\\   
    \textbf{EK-FAC} \cite{Studying_large_language}  & 22.7 &25.6 &29.9 &31.9 & -  
     &31.7 &37.1 &42.7 & 46.0 & -  
      &21.1 &25.8 &31.7 &36.0 & -  
       &44.0 &47.9 &54.8 &60.2 & -  \\
    \textbf{GEX} \cite{GEX}
    & 22.2 & 27.5 & 30.6 & 33.3  &-
    & 31.6 & 37.9 & 42.9 & 47.1 &-
    & 21.9  &26.0  & 35.9 & 39.3  &-
    & 43.1 & 51.4 & 57.1 & 61.6  &- \\    
    \textbf{TRAK}  \cite{Attribution_acale} &23.2 &26.5 &29.3 &33.0 & -  
     &33.8 &38.9 &43.1 &47.2 & -  
      &22.3 &27.5 &36.2 &39.3 & -  
       &47.0 &52.2 &58.4 &\textbf{63.1} & -  \\
    \textbf{TracIn}  \cite{TracIN}
    & \textbf{24.7} & 27.1 & 31.9 & \textbf{34.5}  &-
    & 33.1 & \textbf{39.5} & \textbf{45.5} & \textbf{47.8} &-
    &  {22.4}  &28.1  & {36.7} & \textbf{40.8}  &-
    & \textbf{47.8} & {53.0} & 58.9 & {63.0}  &- \\
    \textbf{DVE-INF} \cite{wang2025capturingDVE}  
    &24.3 &26.4 &29.9 &30.5 & -  
     &33.7 &38.1 &40.9 &44.3 & -  
      &21.9 &27.4 &34.9 &39.7 & -  
       & 46.6 &52.8 &57.1 &62.3 &-  \\ 
    \midrule[0.8pt] 
    \textbf{Diff-In-F}  
    & 24.3 & \textbf{27.8} & \textbf{32.9} & 34.4  &-
    & \textbf{34.2} & \textbf{39.5} & 45.4 & {47.7} &-
    & {23.5}  &\textbf{29.0}  & \textbf{37.1} & 40.7  &-
    & \textbf{47.8} & \textbf{53.1} & \textbf{59.5} & 62.5  &- \\   
    \textbf{Diff-In}  
    & \textbf{25.8} & \textbf{28.3} & \textbf{33.5} & \textbf{36.2}  &-
    & \textbf{34.4} & \textbf{40.6} & \textbf{46.4} & \textbf{48.9} &-
    & \textbf{23.9}  &\textbf{29.7}  & \textbf{38.0} & \textbf{41.1}  &-
    & \textbf{48.3} & \textbf{53.7} & \textbf{60.2} & \textbf{63.4}  &- \\     
\bottomrule
\end{tabular}
}
\vspace{-10pt}
\end{table*}

\subsection{Coreset Selection} 
\label{subsec:exp_coreset_selection}

Coreset selection aims to identify a compact subset of the training data in such a way that the model's performance on this subset closely approximates its performance on the entire dataset \cite{k_center, SSP}. The proportion of the selected subset to the original dataset is termed the selection ratio. We carried out experiments on both an image classification task and a vision-language pretraining task. The detailed experimental setups can be found in the supplementary material. For all experiments, we first pre-train a model on the full training set, then calculate the influence score for each training sample on the training loss \cite{IF, GEX, kwon2023datainf}, that is, $\mathcal{I}(z, \mathbf{D})$. A higher influence score on the training loss indicates that a sample may have a more positive influence \cite{tan2023data}. Finally, we retain the subset corresponding to the highest influence scores as the coreset.

We chose several well-known influence estimators as baselines, including IF \cite{IF}, GEX \cite{GEX}, and DataInf \cite{kwon2023datainf}, which do not account for training-dynamic awareness, as well as TracIn \cite{TracIN} and DVE-INF \cite{wang2025capturingDVE}, which do. In addition to these influence estimators, we also compared Diff-In with leading coreset selection methods, such as Prototypicality (also known as SSP) \cite{SSP} and Moderate \cite{Moderate}, along with a well-known data attribution method, TRAK \cite{Attribution_acale}.

\vspace{0.3cm}
\subsubsection{Experiments on classification datasets} 

We carried out experiments on three public benchmarks: CIFAR-100 \cite{CIFAR}, Tiny-ImageNet \cite{tiny}, and ImageNet-1K \cite{imagenet}. Then, we train a ResNet model \cite{resnet} using the contaminated data (ResNet-18 for CIFAR and Tiny-ImageNet, ResNet-50 for ImageNet-1K). We use the trained model to calculate the influence scores of the samples. The various methods will select different proportions of coreset samples, which will be used to retrain a model. The accuracy of this model on the validation set will be considered as the performance measure for the data selection method.

The experimental results are presented in Table \ref{tb: coreset}. Among all the influence estimators, Diff-In consistently achieves the best results across most settings and datasets. This superiority is especially prominent at lower selection ratios. For instance, on Tiny-ImageNet, Diff-In outperforms the best baseline estimator (TracIn) by 2.4\% at a selection rate of 20\%. 
On ImageNet, Diff-In surpasses TracIn by 3.9\% at a selection rate of 30\%. 
Moreover, even when contrasted with methods specifically designed for coreset selection, such as SSP and Moderate Coreset, Diff-In remains highly competitive, sustaining superior performance in nearly all settings—except for the 70\% setting on CIFAR-100 and Tiny-ImageNet and the 40\% setting on ImageNet-1K.

\begin{table*}[tp]
\setlength{\tabcolsep}{3.1pt}
\caption{\label{tb: Further ablation study for m on CC12M}
Further ablation study on the settings of the number of selected time steps for coreset selection in CC12M \cite{CC12M} (using the influence on training loss as that in Sec. \ref{sec: coreset clip}) and data cleaning in GSM8K \cite{cobbe2021gsm8k} (using the self-influence as that in Sec. \ref{sec: experiment cleaning}).
}
\centering
\resizebox{0.8508\linewidth}{!}{ 
\begin{tabular}{cccccccl}
\toprule
\multirow{2}{*}{\textbf{The choice of $m$ $(\downarrow)$}}  & \multicolumn{4}{c}{\textbf{Coreset selection on CC12M} \cite{CC12M}} & \multirow{2}{*}{\textbf{Data Cleaning on GSM8K} \cite{cobbe2021gsm8k}}\\
\cmidrule(lr){2-5}
~ & $~~$ {\textbf{Zero-shot classification}} $~~$ &  {\textbf{Linear Prob}} $~~$ &  {\textbf{I2T Retrieval}} $~~$ &  {\textbf{T2I Retrieval}} $~~$ \\
\midrule
\textbf{3} & $24.3$ & $33.2$ & $20.7$ & $46.5$ & 91.9 \\
\textbf{5} & $25.8$ & $34.4$ & $23.9$ & $48.3$ & 92.2 \\
\textbf{10} & $25.9$ & $34.4$ & $24.2$ & $48.2$ & 92.4 \\
\textbf{20} & $26.1$ & $34.6$ & $24.4$ & $48.2$ & 92.3 \\
\textbf{30} &26.3 &34.6 &24.5 &48.3 & 92.2 \\
\textbf{40} &26.4 &34.6 &24.5 &48.2 & 92.1 \\
\textbf{50} &26.5 &34.7 &24.5 &48.3 & 92.2 \\
\bottomrule
\end{tabular} 
}
\end{table*}

\vspace{0.3cm}
\subsubsection{Experiments on large models} 
\label{sec: coreset clip}

We conducted large-scale experimental comparisons on the vision-language pretraining task. The dataset used is the CC12M \cite{CC12M}, which consists of 12 million image-text pairs sourced from the Internet for CLIP-like vision-language pre-training. The pre-trained CLIP model \cite{CLIP} is frequently employed to rate image-text data from the vision-language dataset, wherein a higher CLIP score denotes superior image-text alignment. Consequently, there exists a prevalent and valuable coreset baseline \cite{LAION, li2022blip, li2023blip2} for vision-language datasets that retain data with higher CLIP scores. In addition to those influence-based approaches, we consider an additional advanced coreset baseline, the \textbf{Moderate} coreset \cite{Moderate}. For all methods, we train a CLIP-Vit-Base model on CC12M; the detailed settings are introduced in the supplementary material. Regarding Diff-In, we use two time-steps: uniformly selecting times (designated as Diff-In) and selecting only the final time step (designated as Diff-In-F). After data selection, we conduct CLIP pre-training on the coreset and evaluate the model on four downstream tasks: Zero-shot ImageNet Classification, Image-to-Text Retrieval, and Text-to-Image Retrieval on Flickr30K \cite{flickr30k}, and Linear Probing on ImageNet; see supplementary material for more details.

The experimental results are shown in Table \ref{tb: coreset clip}. 
For all 12M samples, it takes about 9.4 minutes to calculate CLIP scores, 10.2 minutes to calculate Diff-In-F, and 59.2 minutes to calculate Diff-In. 
Diff-In consistently outperforms all baselines across all selection ratios, surpassing the CLIP score by 1\%-2\% in most cases. 
Notably, with a similar time cost as the popular CLIP score, Diff-In-F performs better than the CLIP score on all experiments except the (selection-rate=10\%) T2I retrieval experiment. 
This is because CLIP scores focus solely on evaluating text-image alignment and favor simpler, more obvious matches, potentially overlooking more complex but informative samples. By considering influence, Diff-In offers a more comprehensive evaluation, considering a sample's alignment and informativeness.

\vspace{0.3cm}
\subsubsection{Generalization test} To test whether the selected coresets are overfitting to the specific network architecture, we assessed their ability to generalize to different architectures using the CIFAR-100 and Tiny-ImageNet datasets in our experiments. Specifically, we evaluated the performance of the coresets on two different architectures: SENet \cite{SENet} and EfficientNet-B0 \cite{ENet}, which were not used in the initial selection process. 

\begin{table}
\setlength{\tabcolsep}{3.1pt}
\caption{\label{tab:generalization}
Generalization to unseen architecture test on CIFAR-100 and Tiny-ImageNet (denoted by Tiny in the table). 
Here, we tested the generalization performance of the coresets selected using various methods on network architectures different from the surrogate network's architecture. We chose two different architectures, SENet \cite{SENet} and EfficientNet-B0 \cite{ENet} (denoted by EB0 in the table). 
In all experiments, the selection ratio of the coresets from all methods is $20\%$. Here we use the influence on training loss as the selection metric as used in Sec.\ref{subsec:exp_coreset_selection}. 
}
\centering
\resizebox{0.999\linewidth}{!}{ 
\begin{tabular}{lccccccl}
\toprule
 \textbf{Settings ($\rightarrow$)} &$~~$  \textbf{SENet CIFAR} &$~~$ \textbf{EB0 CIFAR}  &$~~$  \textbf{SENet Tiny} &$~~$ \textbf{EB0 Tiny} $~~$ \\
\midrule
\textbf{Random} & $53.57$                              & $42.42$       &$34.13$      &$32.88$  \\  
\textbf{SSP} \cite{SSP} & $54.16$                  & $43.65$        &$31.74$       &$30.99$    \\
\textbf{Moderate} \cite{Moderate}  & {55.57}  & {48.58}       &{36.04}        &{34.26}  \\
\midrule 
\textbf{IF} \cite{IF}  &37.81 &36.94 &32.82 &30.75 \\
\textbf{DataInf} \cite{kwon2023datainf}  &38.98 &39.42 &33.24 &31.63\\
\textbf{GEX} \cite{GEX} &51.67 &42.87 &34.84 &31.93 \\
\textbf{TracIn} \cite{TracIN} &53.12 &46.09 &34.25 &33.54 \\
\textbf{EK-FAC} \cite{Studying_large_language} &49.93 &38.51 &32.22 &30.34  \\
\textbf{TRAK}\cite{Attribution_acale} &52.65 &43.48 &34.31 &32.69  \\
\textbf{DVE-INF} \cite{wang2025capturingDVE}  &53.58 &46.59 &34.99 &33.86  \\ 
\midrule
\textbf{Diff-In}  &\textbf{55.62}     &  \textbf{48.82}    & \textbf{36.41}     & \textbf{34.52}\\
\bottomrule
\end{tabular} 
}
\end{table}

Based on the results presented in Table \ref{tab:generalization}, our approach consistently outperforms other methods in terms of generalization to unseen architectures. Specifically, our method achieves the highest performance on both SENet and EfficientNet-B0 architectures for both CIFAR-100 and Tiny-ImageNet datasets. 
For example, on the EfficientNet-B0 architecture over CIFAR-100, our method achieves an accuracy of 48.82\%, which is 0.24\% higher than the second-best method (Moderate). 
On the Tiny-ImageNet dataset, our method again achieves the highest accuracy of 36.41\% and 34.52\% on the SENet and EfficientNet-B0 architectures, respectively. 
It is worth noting that some methods, such as IF, DataInf, and GEX, sometimes perform even worse than the randomly selected random coreset, while our Diff-In achieves the largest margin of improvement over random selection. This demonstrates the good generalization ability of our method.

\subsection{Further discussion}  
\label{subsec: Preliminary experiments}

Here, we conduct studies on the approximation precision, speed tests, and the effect of $m$ (the number of sampled time steps). 

\vspace{0.3cm}
\subsubsection{The effect of the choice of $m$}

Here, we present additional studies on the choice of the hyperparameter $m$ (which represents the number of selected time steps in the calculation of Diff-In) and provide guidelines for selecting random time steps. This study was performed on two large-scale scenarios: coreset selection in vision-language pre-training (VLP) using CLIP-ViT-Base on CC12M \cite{CC12M}, and data cleaning for LLM-SFT dataset on GSM8K \cite{cobbe2021gsm8k}. 

First, we present the VLP results at a selection rate of 10\% in Table \ref{tb: Further ablation study for m on CC12M}. The setups are consistent with those in Sec.\ref{sec: coreset clip}. The results indicate that as the selection rate increases, both performance and stability improve. However, this improvement plateaus once it exceeds 5. Next, on GSM8K \cite{cobbe2021gsm8k}, we varied the choice of \( m \) from 3 to 50 for our Diff-In method with the selection rate as 30\% and the noise rate as 20\%. The results, presented in Table \ref{tb: Further ablation study for m on CC12M}, exhibit the same trend as observed in the coreset selection for CC12M. As \( m \) increases, both performance and stability improve, with diminishing returns occurring once \( m \geq 5 \). At this point, increasing \( m \) leads to a significant increase in processing time, which is not cost-effective. Therefore, we recommend adhering to a maximum \( m = 5 \) configuration for Diff-In in general scenarios. In time-sensitive situations, we even suggest using only the final checkpoint, referencing the performance results for Diff-In-F in Table \ref{tb: coreset clip} and the time results in Table \ref{tb: Further study on the speed}.

\vspace{0.3cm}
\subsubsection{Approximation precision comparison} 

We conducted a comparative analysis of the approximation accuracy of Diff-In against various other influence estimators. 
We choose 30 data points with the highest influence scores for each type of influence estimator and then calculate the \textbf{Pearson-correlation} between the estimated values and the exact value obtained by the brute-force LOO retraining. 
The experiment is conducted on the CIFAR-100 dataset \cite{CIFAR} using ResNet-18 \cite{resnet}. 
We also considered another counterfactual-type metric commonly used in data attribution, the \textbf{Linear Data Modeling Score (LDS)}. This metric measures the impact on the model of removing samples from a random group, as compared to the total score obtained by summing the data attribution scores for that group, assessed through Spearman correlation. In the experiments described above, we focused on validation influence. For baselines, we compare our approach with Koh's method \cite{IF} (abbreviated as IF), the accelerated version EK-FAC \cite{Studying_large_language}, TRAK \cite{Attribution_acale}, and TracIn \cite{TracIN}. Additionally, due to the extremely slow computation speed of SGD-Inf \cite{hara2019data} (which takes hundreds to thousands of seconds to process a single data point, a time cost we cannot afford), we opted for the latest DVE-IF \cite{wang2025capturingDVE}, which is an accelerated successor to SGD-Inf. For specific results, please refer to Table \ref{tb: approximation accuracy test}. 

Diff-In achieves the best performance across both selected metrics! In terms of the \textbf{Pearson correlation} metric, it significantly outperforms the second-best DVE-IF by approximately 0.1. Similarly, Diff-In also ranks highest in the \textbf{LDS score}, outperforming the second-place TRAK \cite{Attribution_acale}. Furthermore, we found that this advantage is maintained under both SGD and Adam optimizer settings. This strongly indicates that Diff-In is a more accurate data attribution method.

\begin{table}[tp]
\setlength{\tabcolsep}{3.1pt}
\caption{\label{tb: approximation accuracy test}
Experimental results on approximation precision (using the influence on validation loss). In this study, the model was trained for 20 epochs. For our Diff-In method, we selected \( m = 5 \), meaning we uniformly chose 5 checkpoints from the training process. It is important to note that TracIn \cite{TracIN} and TRAK \cite{Attribution_acale} do not estimate the influence of leave-one-out (LOO) retraining; therefore, we have not reported their correlation with the true LOO results. 
}
\centering
\resizebox{0.99\linewidth}{!}{ 
\begin{tabular}{lcccccccl}
\toprule
  &  \multicolumn{2}{c}{\textbf{SGD-Optimizer}} &  \multicolumn{2}{c}{\textbf{Adam-Optimizer}} \\
  \midrule
 \textbf{Method ($\downarrow$)} &  \textbf{Pearson-corr} & \textbf{LDS score} \cite{Attribution_acale} &  \textbf{Pearson-corr} & \textbf{LDS score} \cite{Attribution_acale}\\
\midrule
\textbf{IF}\cite{IF} &0.328 &0.009 &0.357 &0.006 \\
\textbf{EK-FAC} \cite{Studying_large_language} &0.354 &0.042 &0.389 &0.051 \\
\textbf{TRAK} \cite{Attribution_acale} & - & 0.240 & - & 0.237 \\
\textbf{TracIn} \cite{TracIN} & - & 0.187 & - & 0.172 \\
\textbf{SGD-INF} \cite{hara2019data} & 0.665 & 0.241 & 0.681 & 0.233 \\
\textbf{DVE-IF} \cite{wang2025capturingDVE} & 0.629 & 0.214 & 0.633 & 0.206 \\
\midrule
\textbf{Diff-In} &\textbf{0.752} &\textbf{0.271}  &\textbf{0.727} &\textbf{0.264}\\
\bottomrule
\end{tabular} 
}
\end{table}

\vspace{0.3cm}
\subsubsection{Speed comparison}

To thoroughly reflect the comparative speed of various methods, we report the total time required for each method to score the 12 million data points in CC12M. In this analysis, we consider settings that utilize only the parameters from the last layer \cite{OPT,tan2023data} of the CLIP vision encoder. For both Diff-In and TracIn, we also set \( m = 5 \), meaning we uniformly selected 5 checkpoints from the training process. The selected baselines include the extended work of IF, namely {EK-FAC} \cite{Studying_large_language}, {TRAK} \cite{Studying_large_language}, {TracIn}, and {DVE-IF} \cite{wang2025capturingDVE}. Regarding Diff-In, we contemplate two time-step configurations: uniformly selecting time steps (designated as Diff-In, the normal setting) and selecting only the final time step (designated as Diff-In-F, the efficient setting).

\begin{table}[tp]
\setlength{\tabcolsep}{3.1pt}
\caption{\label{tb: Further study on the speed}
Comparison of the computational speed on CC12M \cite{CC12M} (using the influence on training loss in Sec.\ref{subsec:exp_coreset_selection}). The used model is a CLIP-ViT-Base model trained on CC12M.  } 
\centering
\resizebox{0.670056\linewidth}{!}{ 
\begin{tabular}{lccccccl}
\toprule
 \textbf{Method ($\downarrow$)} &  $~~~~~$\textbf{Overall Time Cost (Hours)}\\
\midrule
\textbf{EK-FAC} \cite{Studying_large_language} &4.14 \\
\textbf{TRAK} \cite{Attribution_acale} &3.59\\
\textbf{TracIn} \cite{TracIN} & 15.94 \\
\textbf{DVE-IF} \cite{wang2025capturingDVE} & 16.05 \\
\midrule
\textbf{Diff-In} & 16.22 \\
\textbf{Diff-In-F} & \textbf{3.22} \\
\bottomrule
\end{tabular} 
}
\end{table}

Experimental results could be found in Table \ref{tb: Further study on the speed}. Note that the speed of Diff-In is not significantly slower than TracIn. This is due to Diff-In's highly efficient approximation methods when computing the Hessian-gradient product, as described in Eq.\eqref{eq: fast hessian}. However, Diff-In-F achieves the fastest speed! Reviewing Table \ref{tb: coreset clip}, we see that although Diff-In-F lags behind Diff-In in performance, it still outperforms the majority of the baselines. Therefore, we recommend using Diff-In-F in time-sensitive scenarios, which relies solely on the final checkpoint from training; otherwise, we suggest using Diff-In.

\section{Conclusion} 

This paper presents a novel formulation, Diff-In, which approximates a sample's influence by accumulating the differences in influence between consecutive learning steps. By using second-order approximations, Diff-In has achieved high accuracy in approximating the difference terms without the need for model convexity required by traditional methods. Despite its second-order nature, it maintains good scalability and efficiency, achieved by computing the Hessian-gradient product with finite differences of gradients. The theoretical analysis and extensive experiments fully indicate its superiority.

\section{Acknowledgements} 

This work has been supported in part by the Hong Kong Research Grant Council - Early Career Scheme (Grant No. 27209621), General Research Fund Scheme (Grant No. 17202422, 17212923), Theme-based Research (Grant No. T45-701/22-R), and the Shenzhen Science and Technology Innovation Commission (SGDX20220530111405040).  Part of the described research work is conducted in the JC STEM Lab of Robotics for Soft Materials funded by The Hong Kong Jockey Club Charities Trust. 

	\bibliographystyle{IEEEtran}
\small{
\bibliography{ref}
}

\begin{IEEEbiography}[{\includegraphics[width=1in,height=1.25in,clip,keepaspectratio]{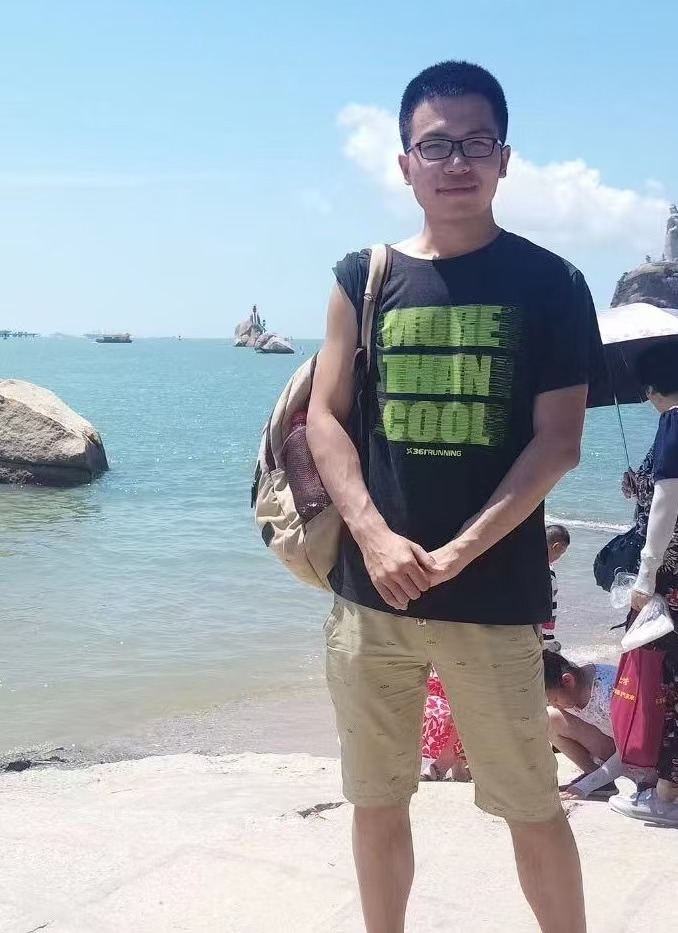}}]{Haoru Tan} is currently a second-year Ph.D student at the University of Hong Kong. He received his master's degree from the Chinese Academy of Sciences, Institute of Automation. He has also had internship experiences at institutions such as Baidu Research, Alibaba DAMO Academy, and Tencent Research. His research interests include data-centric AI, machine learning, and optimizations. He has published more than 10 papers at top journals and conferences, including NeurIPS, ICLR, ICCV, CVPR, IJCV, and T-PAMI. 
\end{IEEEbiography}

\begin{IEEEbiography}[{\includegraphics[width=1in,height=1in,clip,keepaspectratio]{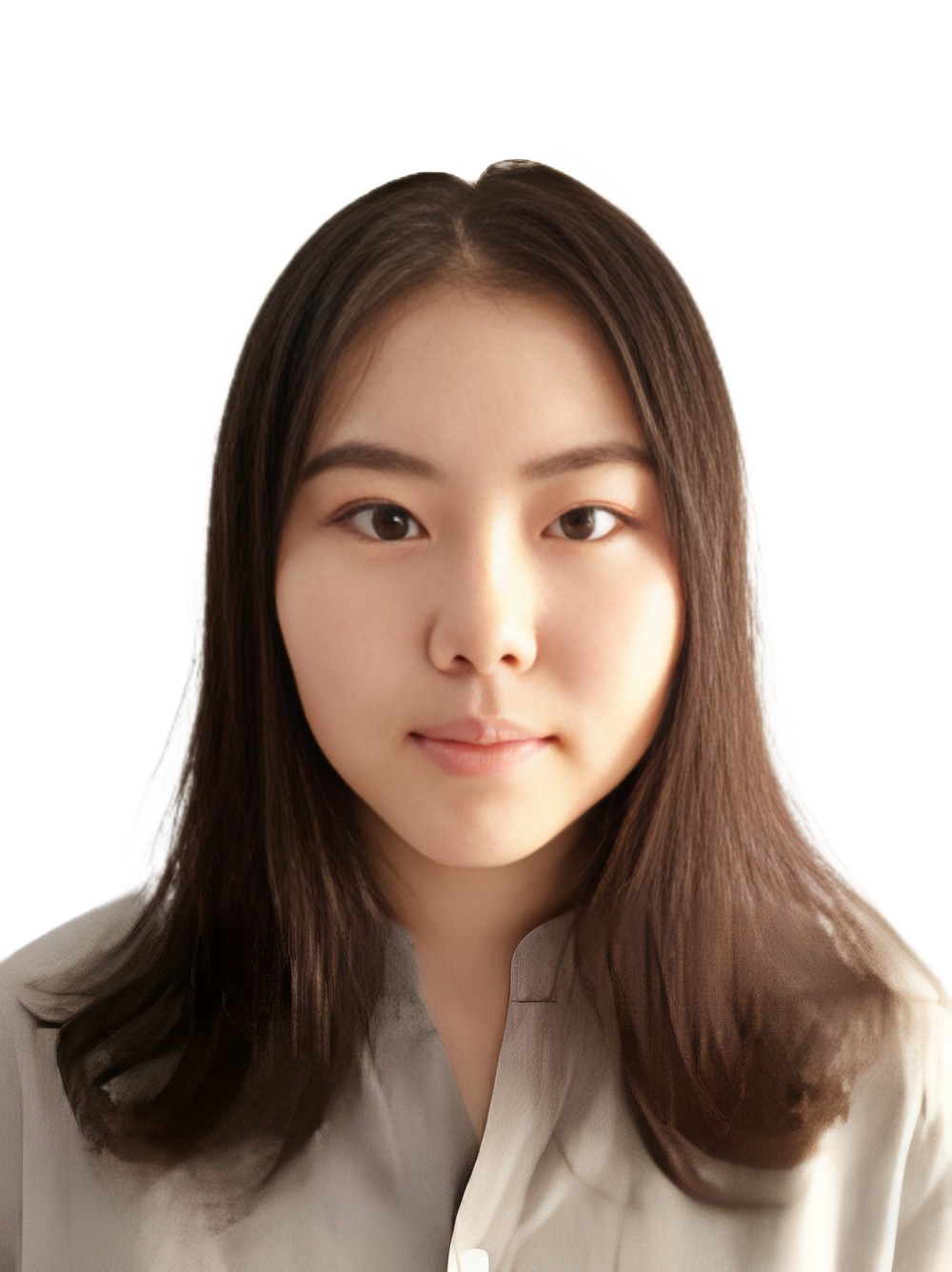}}]{Sitong Wu} is currently working towards the Ph.D. degree with the Department of Computer Science and Engineering, the Chinese University of Hong Kong. She has also had internship experiences at institutions such as Baidu Research and Alibaba DAMO Academy. 
Her current research interests include efficient AI and data-centric AI, particularly exploring their applications in current hot topics, such as large language models, multi-modality models, and generative models. She has published more than 10 papers at top journals and conferences, including ICCV, CVPR, NeurIPS, ICLR, AAAI, etc. 
\end{IEEEbiography}

\begin{IEEEbiography}[{\includegraphics[width=1in,height=1in,clip,keepaspectratio]{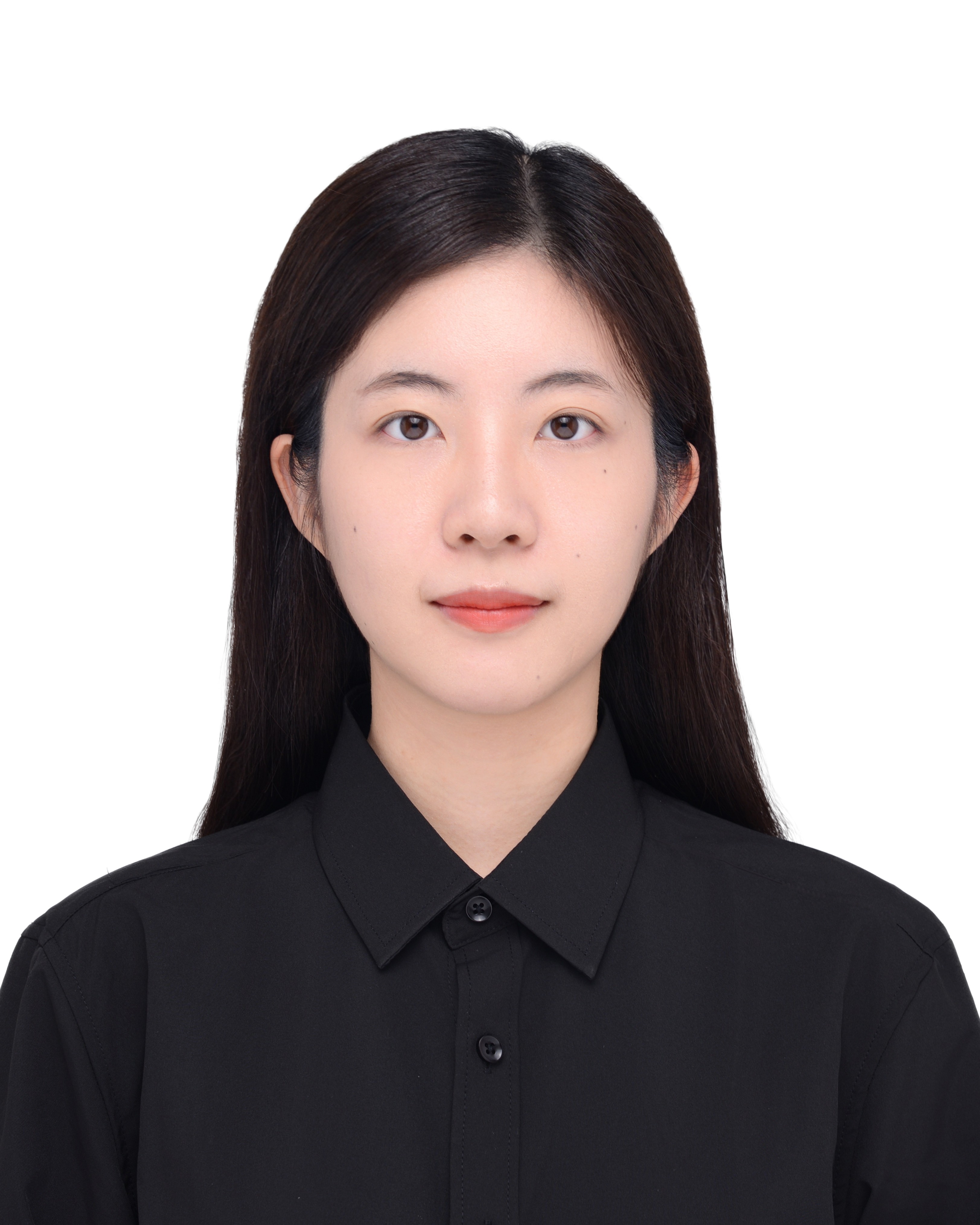}}]{Xiuzhe Wu} received her B.S. and M.S. degrees in Computer Science from Tongji University, and her Ph.D. in Electrical and Electronic Engineering from the University of Hong Kong. She is currently a postdoctoral researcher at the School of Medicine, Stanford University. Her research interests include image/video processing, computer vision, and medical vision-language models. 
\end{IEEEbiography}

\begin{IEEEbiography}[{\includegraphics[width=1in,height=1in,clip,keepaspectratio]{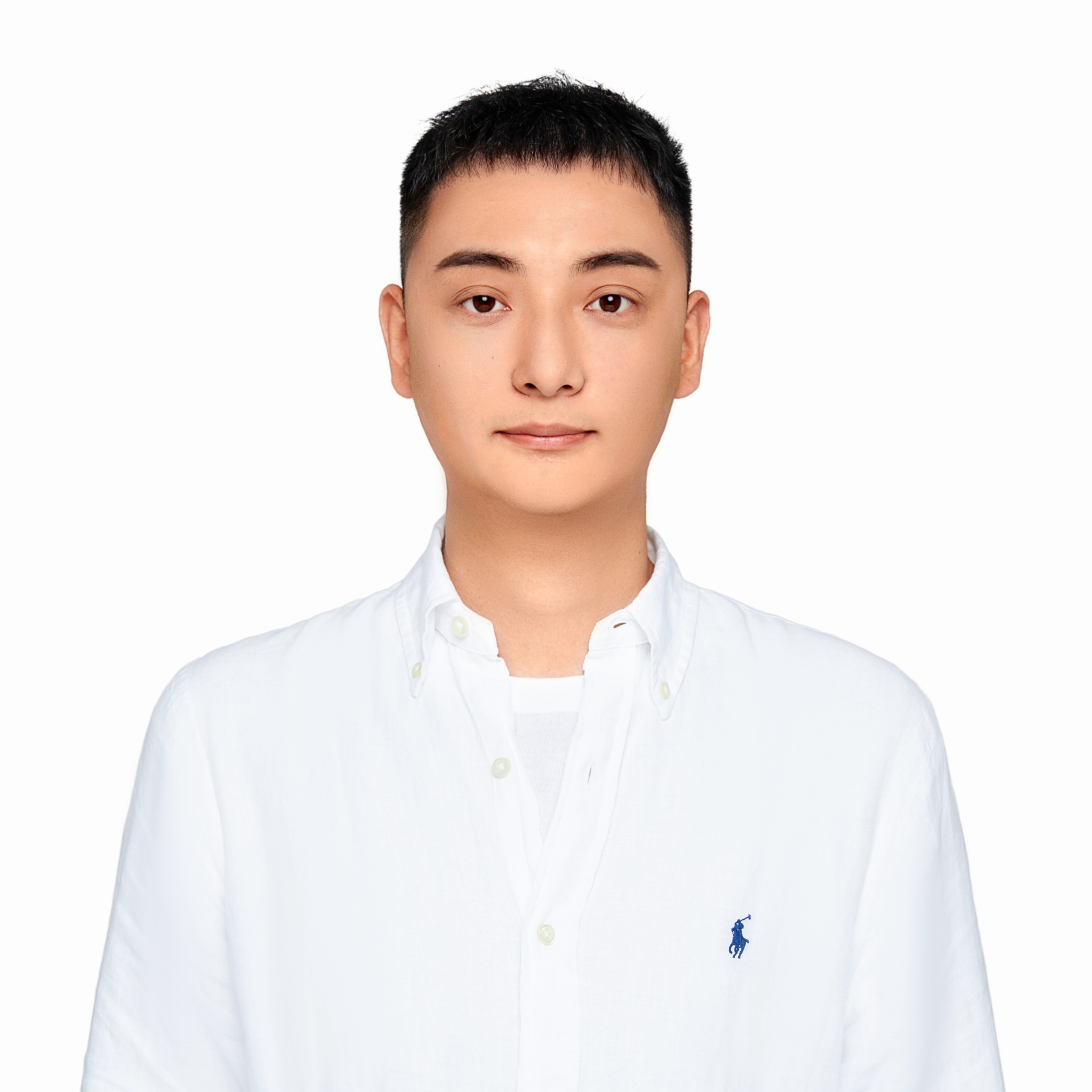}}]{Wang Wang} is an incoming Ph.D. student at The University of Hong Kong. His research interests lie at the intersection of computer vision and engineering applications. He is dedicated to applying computer vision techniques to large-scale civil infrastructure, with a particular focus on structural health monitoring (SHM) and the development of digital twins for bridge structures.
\end{IEEEbiography} 

\begin{IEEEbiography}[{\includegraphics[width=1.2in,height=1.2in,clip,keepaspectratio]{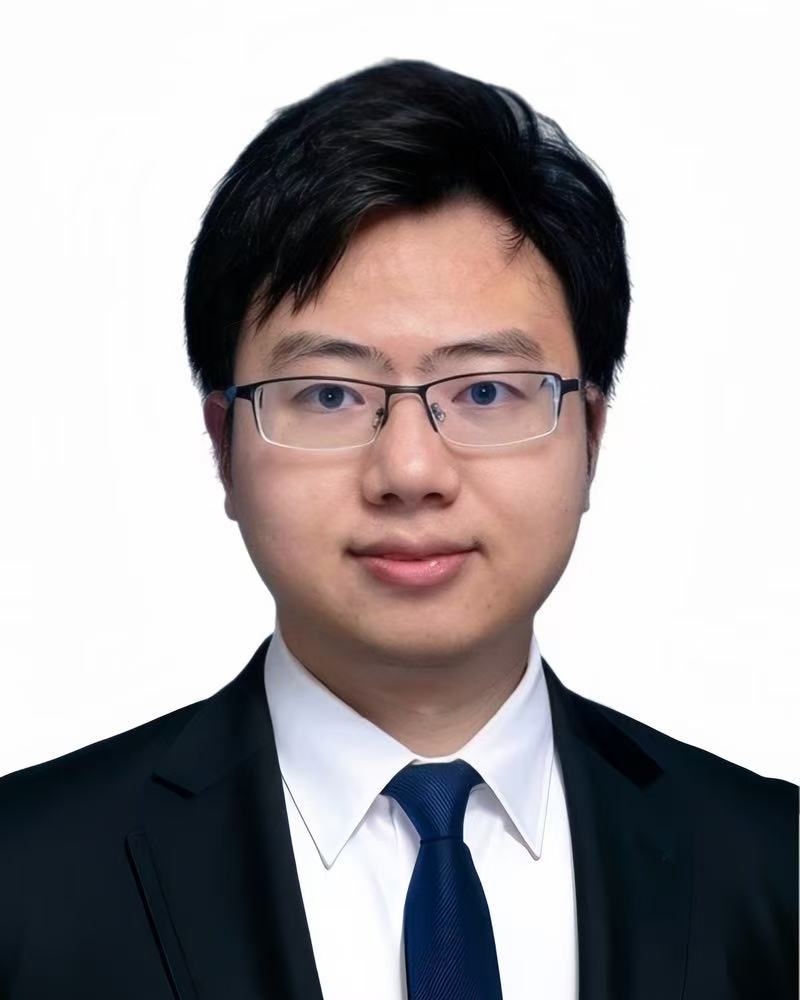}}]{Dr. Zeke Xie} is an Assistant Professor at Information Hub, Hong Kong University of Science and Technology (Guangzhou).  He is leading the xLeaF Lab that is generally interested in understanding and solving fundamental issues of modern AI, particularly AIGC and Large Models, by scientific principles and methodology. He received multiple competitive faculty research awards from ByteDance, Huawei, and CCF-Baidu. He also served as Area Chairs for top conferences, including NeurIPS.
\end{IEEEbiography}

\begin{IEEEbiography}[{\includegraphics[width=1.2in,height=1.2in,clip,keepaspectratio]{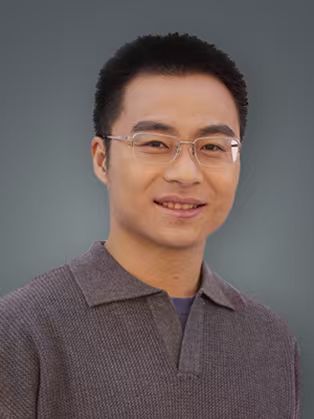}}]{Bo Zhao} is an Associate Professor from School of Artificial Intelligence, Shanghai Jiao Tong University. Before, he was with BAAI, leading the Data-centric AI research team. He received Ph.D. degree from The University of Edinburgh. He is working on Data-centric AI, Multimodal LLM and Embodied AI. He has published dozens of top conference/journal papers, including ICLR Oral, CVPR Oral papers, NeurIPS Spotlight. He served as area chair for NeurIPS and BMVC. He received ICML 2022 Outstanding Paper Award.
\end{IEEEbiography}

\begin{IEEEbiography}[{\includegraphics[width=1.2in,height=1.2in,clip,keepaspectratio]{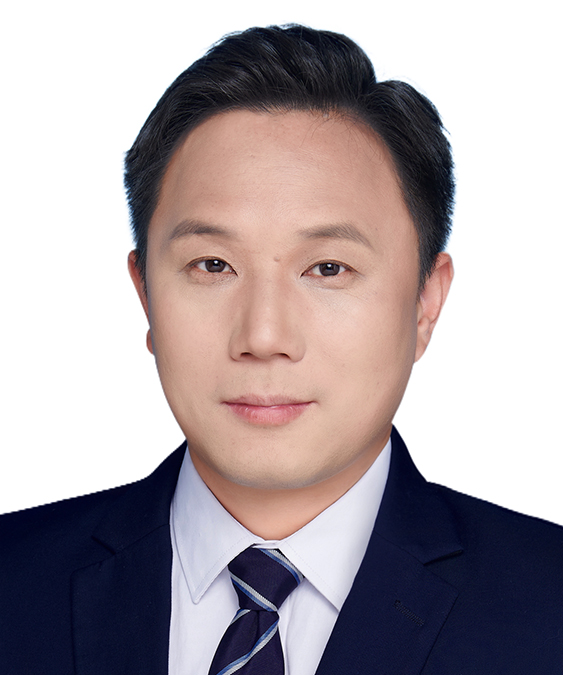}}]{Gui-Song Xia} (Senior Member, IEEE) received the PhD degree in image processing and computer vision from CNRS LTCl, Télécom Paris, Paris, France, in 2011. From 2011 to 2012, he has been a postdoctoral researcher with the Centre de Recherche en Mathématiques de la Decision, CNRS, Paris-Dauphine University, Paris, for one and a half years. He is currently a Hongyi distinguished professor with Wuhan University, leading a research group working on computer vision and robotics. He has also been a visiting scholar with DMA, Ecole Normale Supérieure (ENS-Paris) for two months in 2018. His current research interests include computer vision, robotics, and remote sensing imaging. He was/is on the Editorial Boards of several journals, including \textit{ISPRS Journal of Photogrammetry and Remote Sensing, Pattern Recognition, Signal Processing: Image Communications, EURASIP Journal on Image \& Video Processing, Journal of Remote Sensing, and Frontiers in Computer Science: Computer Vision}. 
\end{IEEEbiography}

\begin{IEEEbiography}[{\includegraphics[width=1in,height=1.25in,clip,keepaspectratio]{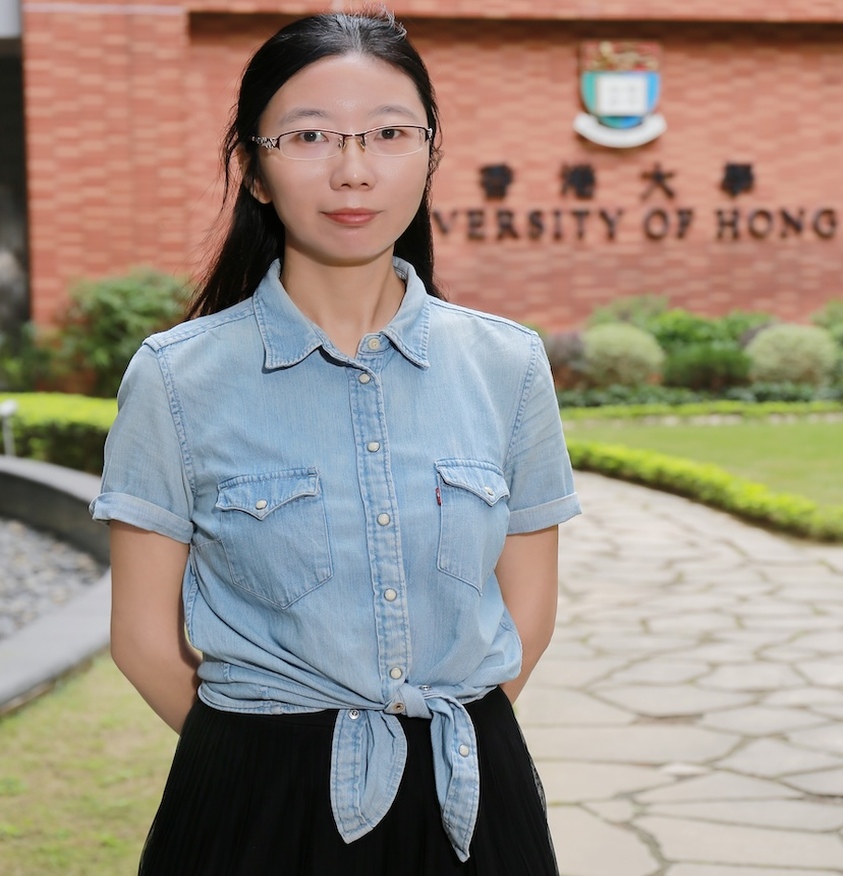}}]{Xiaojuan Qi}
(Senior Member, IEEE) is currently an assistant professor at the Department of Electrical and Electronic Engineering (EEE) of the University of Hong Kong, leading the Computer Vision and Machine Intelligence Lab (CVMI Lab). Before that, she spent 1 year at the University of Oxford, UK, as a postdoctoral scholar with Prof. Philip Torr, and obtained her Ph.D. from The Chinese University of Hong Kong, Hong Kong, and B.S. from Shanghai Jiao Tong University, China. Her research encompasses the broad areas of Computer Vision, Deep Learning, and Artificial Intelligence. She has published more than 100 cutting-edge papers and has been cited over 38000 times on Google Scholar, and has also been named among 'IEEE AI's 10 to Watch for 2024' and '35 Innovators Under 35 for China by MIT Technology. She has served as the area chair of NeurIPS, ICCV, CVPR, ECCV, and other top conferences many times. 
\end{IEEEbiography}

\vfill

\end{document}